\documentclass[conference]{IEEEtran}
\IEEEoverridecommandlockouts
\usepackage{tabularx}
\usepackage{booktabs}
\usepackage{bbm, graphicx}
\usepackage{subfigure, float}
\usepackage{cite}
\usepackage{multirow}
\usepackage{xcolor}
\usepackage{amsmath,amsfonts,amssymb}
\usepackage{array}
\usepackage{balance}

\usepackage{stackengine}

\usepackage{color}
\usepackage{bm}

\newcommand{\PreserveBackslash}[1]{\let\temp=\\#1\let\\=\temp}
\newcolumntype{C}[1]{>{\PreserveBackslash\centering}p{#1}}
\newcolumntype{R}[1]{>{\PreserveBackslash\raggedleft}p{#1}}
\newcolumntype{L}[1]{>{\PreserveBackslash\raggedright}p{#1}}

\definecolor{ballblue}{rgb}{0.004, 0.50, 0.69}
\usepackage[colorlinks]{hyperref}
\AtBeginDocument{%
  \hypersetup{
    citecolor=ballblue,
    linkcolor=ballblue,   
    urlcolor=ballblue}}

\def\BibTeX{{\rm B\kern-.05em{\sc i\kern-.025em b}\kern-.08em
    T\kern-.1667em\lower.7ex\hbox{E}\kern-.125emX}}
\begin{document}

\title{Focus on Focus: Focus-oriented Representation Learning and Multi-view Cross-modal Alignment for Glioma Grading}



\author{\IEEEauthorblockN{Li Pan\textsuperscript{1*}, Yupei Zhang\textsuperscript{1*}, Qiushi Yang\textsuperscript{2}, Tan Li\textsuperscript{3}, Xiaohan Xing\textsuperscript{4}, Maximus C.\,F. Yeung\textsuperscript{1}, Zhen Chen\textsuperscript{5\dag}}
\IEEEauthorblockA{
\textsuperscript{1}\textit{The University of Hong Kong} \,\, \textsuperscript{2}\textit{City University of Hong Kong} \,\, \textsuperscript{3}\textit{The Hang Seng University of Hong Kong} \\
\textsuperscript{4}\textit{Stanford University} \,\, \textsuperscript{5}\textit{Centre for Artificial Intelligence and Robotics (CAIR), HKISI-CAS}}
\thanks{L. Pan and Y. Zhang contribute equally to this work.}
}


\maketitle

\begin{abstract}
Recently, multimodal deep learning, which integrates histopathology slides and molecular biomarkers, has achieved a promising performance in glioma grading. Despite great progress, due to the intra-modality complexity and inter-modality heterogeneity, existing studies suffer from inadequate histopathology representation learning and inefficient molecular-pathology knowledge alignment. These two issues hinder existing methods to precisely interpret diagnostic molecular-pathology features, thereby limiting their grading performance. Moreover, the real-world applicability of existing multimodal approaches is significantly restricted as molecular biomarkers are not always available during clinical deployment.  To address these problems, we introduce a novel Focus on Focus (FoF) framework with paired pathology-genomic training and applicable pathology-only inference, enhancing molecular-pathology representation effectively. Specifically, we propose a Focus-oriented Representation Learning (FRL) module to encourage the model to identify regions positively or negatively related to glioma grading and guide it to focus on the diagnostic areas with a consistency constraint. To effectively link the molecular biomarkers to morphological features, we propose a Multi-view Cross-modal Alignment (MCA) module that projects histopathology representations into molecular subspaces, aligning morphological features with corresponding molecular biomarker status by supervised contrastive learning. Experiments on the TCGA GBM-LGG dataset demonstrate that our FoF framework significantly improves the glioma grading. Remarkably, our FoF achieves superior performance using only histopathology slides compared to existing multimodal methods. The source code is available at \url{https://github.com/peterlipan/FoF}. 
\end{abstract}


\section{Introduction}\label{sec1}
As the most common type of brain tumors, gliomas are classified into Grade II to IV by the World Health Organization, correlating with varied prognoses and intervention approaches \cite{louis20212021}. The gold standard for grading gliomas is the observation of representative histopathology features in biopsies \cite{louis20162016}. 
However, histopathology slides present a complex milieu of cells, necrosis, and microenvironments, which complicates the localization of tumor foci, necessitating the expertise of senior pathologists \cite{shanes1987interobserver}. The recent advances in computer-assisted cancer grading reveal promising performance in identifying glioma grades from histopathology slides \cite{anand2020histographs, wang2020weakly}. 

Recent studies \cite{chen2020pathomic, mobadersany2018predicting} in cancer grade classification benefit from multimodal approaches that combine histopathology features with molecular biomarkers from tissue biopsies, offering a comprehensive and accurate tumor analysis. Existing multimodal methods \cite{chen2020pathomic, alwazzan2024foaa} heavily rely on paired pathology-genomic data for inference when deployed. However, obtaining molecular biomarkers requires staining and/or sequencing, which is not routinely available in clinical practice due to high costs and technical challenges, thereby limiting their practical applications \cite{zhou2021distilling}. To address this limitation, cutting-edge multimodal techniques \cite{xing2024comprehensive, zhang2023multi} propose achieving uni-modal inference by distilling knowledge from a pathology-genomic teacher, expanding the applicability of multimodal grading in clinical settings. Nevertheless, these methods are inherently constrained by the performance of multimodal teachers, highlighting an emergent need for a novel approach that can effectively learn from multimodal data while enabling unimodal inference.

The effectiveness of existing multimodal methods is limited by the difficulties in accurately identifying diagnostic molecular-pathology features, which can be attributed to two major issues.
The first issue is the \textit{insufficient histopathology representation learning}, attributed to a lack of focus on the most representative regions associated with gliomas. Existing methods \cite{anand2020histographs} attempt to correlate the complex images with diagnostic grades by uniformly processing the whole histopathology slides. Nonetheless, without an explicit regularization, these approaches \cite{wang2020weakly, wang2019machine} are prone to overfitting the minor textual details rather than high-level diagnostic patterns.
Existing studies \cite{chen2020pathomic, zhou2019cgc} attempt to tackle this issue by augmenting the histopathology inputs with handcrafted morphological features but lack the generalization ability. 
To this end, there is an urgent need for a novel method to boost histopathology representation learning by directing the model focus toward areas most relevant to glioma grading. 

Another challenge lies in the \textit{inefficient molecular-pathology knowledge alignment}. Contemporary multimodal methods tend to fuse the pathology-genomic information in the feature space through simple combinations. However, due to the notable heterogeneity between the two modalities, these methods struggle to align the morphological features of tumor niches with molecular biomarkers, such as Intradialytic hypotension (IDH) mutation status and 1p/19q codeletion presence, which are crucial for precise glioma grading \cite{louis20212021, wang2023multi}.
Thus, there is a high demand for novel strategies that can effectively correlate genomic biomarkers with histopathology abnormalities to jointly identify the molecular-pathology features of gliomas toward precision grading.

To tackle these challenges, we propose the Focus on Focus (FoF) framework that enhances the identification of diagnostic molecular-pathology features by boosting the histopathology representations with Focus-oriented Representation Learning (FRL), and effectively integrating the molecular biomarkers with Multi-view Cross-modal Alignment (MCA). 
Specifically, to improve the representations of histopathology slides, we propose FRL to identify the regions positively and negatively correlated to cancer grading and encourage the model to focus on diagnostic regions with a consistency constraint. 
To correlate the genomic biomarkers with histopathology properties, we propose MCA to cluster the multi-view visual representations that share compatible molecular contexts while distancing those in discrepancies through supervised contrastive learning.
By precisely locating the tumor niches and linking them with molecular biomarkers, FoF enriches the visual representations of histopathology slides and facilitates accurate glioma grading only using images. Experimental results prove that our FoF framework outperforms state-of-the-art histopathology and multimodal methods on the TCGA GBM-LGG dataset.

\begin{figure*}[t]
\centering
\makebox[\textwidth][c]{\includegraphics[width=.99\textwidth]{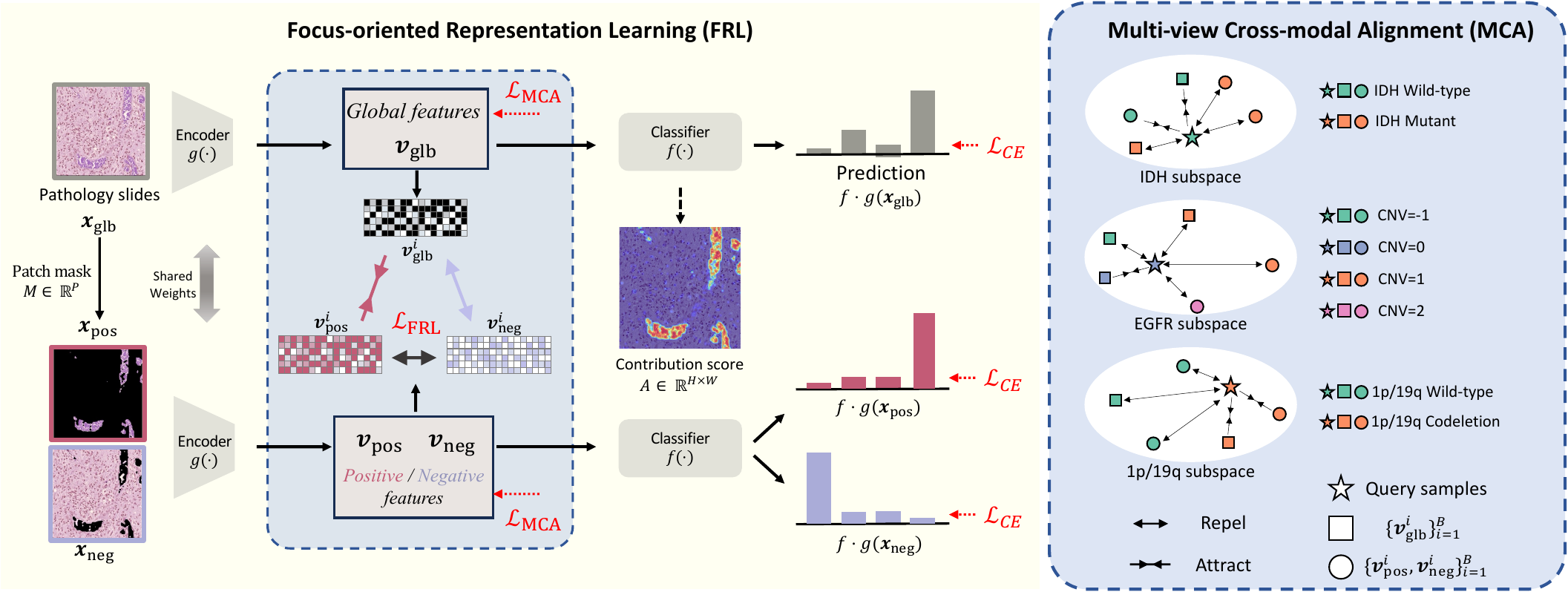}}
\caption{The FoF framework. FRL identifies the positive $\boldsymbol{x}_{\rm{pos}}$ and negative regions $\boldsymbol{x}_{\rm{neg}}$, encouraging the model to focus on the most important regions with a consistency constrict. The MCA module correlates the histopathology representations $\{\boldsymbol{v}_{\rm{glb}}, \boldsymbol{v}_{\rm{pos}}, \boldsymbol{v}_{\rm{neg}} \}$ with molecular biomarkers.} \label{fig:framework}
\end{figure*}

The contributions of this work are summarized as follows:
\begin{itemize}
    \item To achieve accurate grading of gliomas, we propose a novel FoF framework that directs the focus of models to diagnostic regions. To the best of our knowledge, this is the first work to adjust the model focus on areas of both histopathology and molecular-pathology significance.
    
    \item We devise the FRL module to identify regions that positively or negatively contribute to grading, encouraging the model to focus on important diagnostic niches through a consistency constraint on histopathology features.

    \item In response to the latest WHO classification of gliomas, we propose the MCA module to align the histopathology features directly with molecular biomarkers and balance the impact of each biomarker through a novel multi-view contrastive learning. 
    
    \item Extensive experiments on the TCGA GBM-LGG dataset prove the effectiveness, and our FoF outperforms multimodal grading state-of-the-arts using only histopathology slides, demonstrating great clinical significance. 

\end{itemize}

\section{Related work}
\subsection{Glioma Grading}
\noindent\textbf{Histopathology grading} Glioma is a significant type of malignant tumor that occurs in the glial cells of the brain or spinal cord and can be classified into Grades II to IV, each correlating with different prognoses \cite{louis20162016}. Histologic properties observed under a microscope from biopsy samples are the gold standard for tumor assessments in clinical practice, \textit{e.g.}, the presence of necrosis and/or microvascular proliferation leads to the diagnosis of high-grade glioma (HGG) \cite{perry2016histologic}. To reduce the workload of pathologists and promote consistent diagnosis, researchers have utilized traditional machine learning and deep learning models to classify morphological features \cite{xing2024comprehensive}. For example, Ker \textit{et al.} employed a pre-trained convolutional neural network to classify histopathology slides into normal, low-grade glioma (LGG), or HGG categories \cite{ker2019automated}. Rathore \textit{et al.} extracted conventional features (\textit{e.g.}, intensity) and textural features (\textit{e.g.}, gray-level co-occurrence matrix) from specimens, and used a support vector machine for classification \cite{rathore2020glioma}. However, handcrafted features only capture detailed textural information, and deep learning models are prone to overfitting it due to the high complexity of histopathology slides. As a result, these approaches are inefficient at capturing the representative patterns of gliomas, lacking both generalization ability and interpretability.

\noindent\textbf{Multimodal grading} According to the latest WHO classification of tumors of the central nerve system \cite{louis20212021}, integrating molecular biomarkers with histopathology slides provides a more comprehensive and precise analysis of gliomas. Inspired by this medical insight, researchers attempt to project the histopathology and molecular data into a uniform latent space, fusing the features of the two modalities through various combinations, such as concatenation \cite{mobadersany2018predicting, wang2021gpdbn}, Kronecker Product \cite{chen2020pathomic, xing2022discrepancy}, and cross attention \cite{alwazzan2024foaa}.
Yet, molecular biomarkers, such as the status of IDH mutation and 1p/19q codeletion, contain precise information that directly correlates with the grading of gliomas \cite{wang2023multi}. As a result of this heterogeneity between the two modalities, existing methods that rely on uniform projection and late fusion are insufficient at aligning cross-modal information through simple combinations and overlook the correlations among different molecular biomarkers. In contrast, our FoF framework encourages the model to adaptively focus on representative regions and directly align these regions with biomarkers, resulting in more robust representation learning and improved interpretability.

\subsection{Multimodal Glioma Grading with Missing Modality}
Although molecular biomarkers are essential for the clinical assessment of cancers, their acquisition necessitates immunohistochemistry (IHC) staining and DNA/RNA sequencing. These processes are time-consuming and costly, thus resulting in limited accessibility, particularly in underrepresented areas \cite{xing2023gradient}. Fusion-based multimodal methods \cite{chen2020pathomic} necessitate the presence of paired histopathology slides and molecular biomarkers, further restricting their practical application in real-world clinical settings. To alleviate this challenge, DDM-net \cite{qiu2024dual} proposed to reconstruct the features of unavailable modality from the available one with the transformation function learned from paired data. Most recent studies have implemented distillation-based methods to enhance the image-based model by distilling knowledge from a pathology-genomic teacher \cite{xing2024comprehensive}. For instance, Xing \textit{et al.} \cite{xing2022discrepancy} devised a novel distillation framework that effectively transfers knowledge from a multimodal teacher to a uni-modal student, achieving performance comparable to multimodal grading using only histopathology slides. Despite efforts to maximize the efficiency of transferring knowledge from multiple modalities to a single modality, these approaches \cite{qiu2024dual, xing2022discrepancy, xing2024comprehensive} are naturally limited by the patterns learned from pathology-genomic pairs, leading to the aforementioned problem of insufficient cross-modal alignment. Unlike these methods, our FoF enhances the representation learning on histopathology slides directly with the guidance of individual molecular biomarkers, efficiently correlating the representative morphological patterns with molecular properties. 

\section{Methodology}
\subsection{Overview}
As illustrated in Fig.~\ref{fig:framework}, our FoF framework improves the glioma grading by highlighting the diagnostic molecular-pathology features. To enhance the representation learning on histopathology slides, we introduce FRL to identify the positive $\boldsymbol{x}_{\rm{pos}}$ and negative regions $\boldsymbol{x}_{\rm{neg}}$ from the input images $\boldsymbol{x}_{\rm{glb}}$ with the pixel-wise contribution score $\mathbf{A}\in \mathbb{R}^{H\times W}$, encouraging the model $f\cdot g$ to focus on the most important areas. In MCA, we project the multi-view features $\{\boldsymbol{v}^i_{\rm{glb}}, \boldsymbol{v}^i_{\rm{pos}}, \boldsymbol{v}^i_{\rm{neg}}\}_{i=1}^B$ into individual molecular subspaces with distinct projectors $h^n(\cdot)$ and employ biomarkers as labels. During inference, the model predicts the glioma grades $f\cdot g(\boldsymbol{x}_{\rm{glb}})$ with only histopathology slides.

\subsection{Focus-oriented Representation Learning}
Existing representation learning on histopathology slides is inadequate, which can be attributed to the lack of focus on the diagnostic morphological patterns \cite{zhou2019cgc}. To boost histopathology representation learning, we propose the Focus-oriented Representation Learning module to encourage the focus of the model on areas most relevant to gliomas and enhance these representations with a consistency constraint. Specifically, FRL quantifies the contribution score of each pixel towards accurate classification $\mathbf{A} \in \mathbb{R}^{H\times W}$ and identifies the areas that are positively and negatively related to grading using a threshold. This module separately feeds the positive $\boldsymbol{x}_{\rm{pos}}$ and negative $\boldsymbol{x}_{\rm{neg}}$ regions into the ViT encoder $g$ and enhances the visual representations with a consistency constraint on the positive $\boldsymbol{v}_{\rm{pos}}$ and global features $\boldsymbol{v}_{\rm{glb}}$.

Technically, locating the focus of models has been formulated as identifying the input pixels that contribute most significantly to the output \cite{selvaraju2017grad}. To identify the diagnostic regions within a specific slide $\boldsymbol{x}$, FRL initially processes the whole image through the model to obtain the predictions $f\cdot g (\boldsymbol{x})$. It then aggregates the gradient of the prediction for the ground-truth class $s$ across each feature map layer $\boldsymbol{v}^k$, thereby estimating the contribution score of each pixel towards accurate grading as follows:
\begin{equation}
\begin{gathered}
    \alpha_k \ = \ \sum_i\sum_j \frac{\partial s}{\partial \boldsymbol{v}^k_{ij}}, \\ \,\, 
    \mathbf{A} \ = \ \Upsilon(\rm{ReLU}(\sum_k \alpha_k \boldsymbol{v}^k), [H,W]),
\end{gathered}
\end{equation}
where $i$ and $j$ represent the location indices within the feature map $\boldsymbol{v}$, $\alpha_k$ is the weight associated with the $k$-th feature map $\boldsymbol{v}^k$, and $\Upsilon(\cdot, [H,W])$ refers to the reshape and resample function that upscales the feature map to match the input image size. Consequently, the matrix $\mathbf{A}$ indicates the contribution of each pixel toward accurately classifying the input image to the target class. Following this step, we divide the pixel-wise contribution score into $P = \left\lfloor\frac{H}{p}\right\rfloor \times \left\lfloor\frac{W}{p}\right\rfloor$ patches, where $p$ is the patch size. The FRL calculates the average patch-wise contribution score, filtering it by a threshold to generate the mask of patches $M \in \mathbb{R}^{P}$. We apply the patch-wise mask on the patchified image to select only the positive and negative patches as follows:
\begin{equation}
\begin{gathered}
    M_{i,j} \ = \ \frac{1}{p^2} \sum_{m=0}^{p-1} \sum_{n=0}^{p-1} \mathbf{A}_{pi+m, pj+n} > \theta, \\ \,\, \boldsymbol{x}_{\rm{pos}} \ = \ \boldsymbol{x}_{\rm{glb}} \odot M, \,\, \boldsymbol{x}_{\rm{neg}} \ = \ \boldsymbol{x}_{\rm{glb}} \odot (1 - M),
\end{gathered}
\end{equation}
where $\theta$ is a pre-defined threshold for contribution scores set at 0.5 in this study, $\odot$ denotes the element-wise multiplication operator and $x_{\rm{glb}} \in \mathbb{R}^{P\times C}$ corresponds to the patches of the input image. Consequently, $\boldsymbol{x}_{\rm{pos}}$ and $\boldsymbol{x}_{\rm{neg}}$ denote the positive and negative patches, respectively. To promote a consistent grading on global and positive regions as well as distinguish the negative ones, FRL applies the Cross-Entropy loss on all the regions $\{\boldsymbol{x}_{\rm{glb}}, \boldsymbol{x}_{\rm{pos}}, \boldsymbol{x}_{\rm{neg}}\}$ as follows:
\begin{equation}
\begin{gathered}
    \sigma(\boldsymbol{x}, y) \ = \  \mathcal{L}_{\rm{CE}}(f\cdot g(\boldsymbol{x}), y)), \\ \,\,
    \mathcal{L}_{\rm{CLS}} = \frac{1}{B} \sum_{i=1}^B \sigma(\boldsymbol{x}^i_{\rm{glb}}, y^i) + \sigma(\boldsymbol{x}^i_{\rm{pos}}, y^i) + \sigma(\boldsymbol{x}^i_{\rm{neg}}, c),
\end{gathered}
\end{equation}
where $B$ indicates the number of samples in a mini-batch, $\mathcal{L}_{\rm{CE}}$ denotes the Cross-Entropy loss, $y^i$ is the label of the $i$-th sample, and $c$ is a constant label for negative areas indicating background. To further boost the representations of histopathology slides, especially for the positive regions, we propose a consistency constraint that promotes a consistent mapping of positive and overall regions as well as a distinguishable mapping of negative regions.
This approach pulls the positive and global features closer together as well as pushing the negative features far apart by the InfoNCE loss \cite{chen2020simple} defined on the features $\boldsymbol{v} = g(\boldsymbol{x})$ as follows:
\begin{equation}
\begin{gathered}
    \phi(\boldsymbol{v}_i, \boldsymbol{v}_j) \ = \  e^{\frac{{\rm cos}(\boldsymbol{v}_i, \boldsymbol{v}_j)}{\tau}}, \\ \,\,
    \psi(\boldsymbol{v}) = \frac{\phi(\boldsymbol{v}_{\rm{glb}}, \boldsymbol{v}_{\rm{pos}})}{\phi(\boldsymbol{v}_{\rm{glb}}, \boldsymbol{v}_{\rm{pos}}) + \phi(\boldsymbol{v}_{\rm{glb}}, \boldsymbol{v}_{\rm{neg}}) + \phi(\boldsymbol{v}_{\rm{pos}}, \boldsymbol{v}_{\rm{neg}})}, \\ \,\,
    L_{\rm{FRL}} \ = \ - \frac{1}{B} \sum_{i=1}^B \log \psi(\boldsymbol{v}_i),
\end{gathered}
\end{equation}
where $\boldsymbol{v}^i_{\rm{glb}}, \boldsymbol{v}^i_{\rm{pos}}, \boldsymbol{v}^i_{\rm{neg}}$ denote the features extracted from global, positive and negative regions, respectively. The ${\rm cos}(\cdot, \cdot)$ represents the cosine similarity function and $\tau \in \mathbb{R}^+$ is a scalar temperature parameter. By locating the areas that contributed the most significantly to accurate grading and regularizing the representations of different regions with the consistency constraint, FRL encourages the model to focus on diagnostic areas, promoting representation learning on the histopathology slides, thus improving the performance of glioma grading.

\subsection{Multi-view Cross-modal Alignment}
Current multimodal glioma grading methods encode and fuse the histopathology slides with molecular biomarkers \cite{chen2020pathomic}. Notably, unlike histopathology representations, molecular biomarkers usually with integer values, e.g., IDH mutation status, provide direct insights related to glioma grading without necessitating neural network processing \cite{louis20212021}. This heterogeneity between two modalities presents a challenge for existing fusion-based methods, which struggle to align molecular indicators with histologic features in the high-dimensional feature space \cite{wang2023multi}.
To tackle this issue, we propose a Multi-view Cross-modal Alignment module that employs genomic biomarkers as unique labels and aligns the histological feature representations within each molecular subspace by supervised contrastive learning. 

Specifically, MCA leverages the biomarkers with discrete values, including IDH mutation status, 1p/19q codeletion presence, and the Copy Number Variation (CNV), as labels. To accommodate the co-occurrence of biomarkers, this module projects multi-view histopathology representations $\boldsymbol{v} \in \{\boldsymbol{v}_{\rm{glb}}, \boldsymbol{v}_{\rm{pos}}, \boldsymbol{v}_{\rm{neg}} \}$ into individual molecular subspaces, employing distinct projection heads $h^n(\cdot)$ for each. Within these subspaces, it aims to bring features that share the same molecular biomarker values closer together while distancing those that differ, thereby enhancing the concordance of histopathology features regarding each biomarker. The loss of Multi-view Cross-modal Alignment is formulated as follows:
\begin{equation}
\begin{gathered}
    \omega(\boldsymbol{v}, i, j) \ = \ \frac{\phi(\boldsymbol{v}_i, \boldsymbol{v}_j)}{\sum_{k=1}^B \mathbbm{1}_{i\neq k} \cdot \phi(\boldsymbol{v}_i, \boldsymbol{v}_k)}, \\ \,\,
    \mathcal{L}_{\rm{MCA}}^n  = - \frac{1}{B} \sum_{i=1}^B  \sum_{j=1}^B \mathbbm{1}_{i\neq j} \cdot \mathbbm{1}_{y_i^n = y_j^n} \cdot \log \omega(h^n(\boldsymbol{v}), i, j), \\ \,\,
    \mathcal{L}_{\rm{MCA}} \ = \ \sum_{n=1}^N \mathcal{L}_{\rm{MCA}}^n,
\end{gathered}
\end{equation}
where $N$ denotes the number of gene biomarkers, $n$ indicates the index of the biomarker, $\mathbbm{1}$ is an indicator function that returns $1$ if the condition is satisfied and $0$ otherwise, and $y^n$ represents the status of the $n$-th biomarker. We set the gene labels of positive regions to the same as global regions and set the negative gene labels to the normal status (\textit{i.e.}, wildtype for IDH and 1p/19q, CNV equals 0 for other biomarkers). Following recent medical studies \cite{louis20212021, sledzinska2021prognostic}, we include the IDH mutation status, 1p/19q codeletion presence, and CNV of PTEN, EGFR, CARD11, and FGFR2 as molecular biomarkers in this study. By enhancing the histological feature representations with FRL and aligning molecular biomarkers to them, FoF promotes a reliable and accurate grading of gliomas, reaching even better performance with sole images than existing multimodal counterparts.

\begin{table*}[t]
\setlength{\tabcolsep}{12pt}
\centering
\caption{Comparison with \textbf{histopathology grading} on the TCGA GBM-LGG dataset.}
\label{tab:path}
\begin{tabular}{l|cccc}
\toprule
Method & AUC (\%) & AP (\%) & Accuracy (\%) & Kappa (\%) \\
\hline
    Baseline & 90.37 $\pm$ 0.82 & 83.05 $\pm$ 2.39 & 74.36 $\pm$ 2.14 & 61.16 $\pm$ 3.84 \\
    PathCNN \cite{xing2022discrepancy} & 90.46 $\pm$ 1.01 & 82.84 $\pm$ 2.07 & 74.07 $\pm$ 1.81 & 60.59 $\pm$ 3.05 \\
    PathGCN \cite{chen2020pathomic} & 90.25 $\pm$ 1.53 & 82.69 $\pm$ 2.11 & 74.20 $\pm$ 1.49 & 61.27 $\pm$ 2.92 \\
    SwinV2-Tiny \cite{liu2022swin} & 91.56 $\pm$ 1.74 & 83.27 $\pm$ 1.26 & 74.34 $\pm$ 1.56 & 62.35 $\pm$ 2.46 \\
    \hline
    KL div \cite{hinton2015distilling} & 91.67 $\pm$ 0.43 & 85.01 $\pm$ 0.95 & 74.58 $\pm$ 1.15 & 61.39 $\pm$ 2.17 \\
    PKT \cite{passalis2018learning} & 91.52 $\pm$ 0.52 & 84.67 $\pm$ 1.03 & 74.77 $\pm$ 1.05 & 61.71 $\pm$ 1.89 \\
    Feats KL \cite{hu2020knowledge} & 91.70 $\pm$ 0.43 & 85.06 $\pm$ 0.74 & 74.82 $\pm$ 1.66 & 61.76 $\pm$ 2.97 \\
    SP \cite{tung2019similarity} & 91.89 $\pm$ 0.45 & 85.38 $\pm$ 0.54 & 75.60 $\pm$ 1.12 & 62.96 $\pm$ 1.96 \\
    RKD \cite{park2019relational} & 91.67 $\pm$ 0.35 & 85.03 $\pm$ 0.65 & 74.97 $\pm$ 1.06 & 62.00 $\pm$ 2.01 \\
    CRD \cite{tian2019contrastive} & 91.86 $\pm$ 0.68 & 85.33 $\pm$ 1.34 & 75.71 $\pm$ 1.27 & 63.16 $\pm$ 2.04 \\
    SCKD \cite{zhu2021student} & 91.31 $\pm$ 0.34 & 84.44 $\pm$ 0.76 & 73.91 $\pm$ 1.28 & 60.42 $\pm$ 2.32 \\
    HKD \cite{zhou2021distilling} & 91.73 $\pm$ 0.32 & 85.18 $\pm$ 0.59 & 75.21 $\pm$ 0.85 & 62.37 $\pm$ 1.49 \\
    CLAT \cite{xing2024comprehensive} & 92.42 $\pm$ 0.58 & 86.34 $\pm$ 1.23 & 76.47 $\pm$ 0.65 & 64.34 $\pm$ 0.85 \\
    \hline
    FoF \textit{w/o} FRL & 93.20 $\pm$ 0.53 & 88.44 $\pm$ 1.48 & 76.89 $\pm$ 1.60 & 64.31 $\pm$ 2.13 \\
    FoF \textit{w/o} MCA & 93.64 $\pm$ 0.42 & 88.26 $\pm$ 0.93 & 76.55 $\pm$ 2.32 & 63.85 $\pm$ 2.55 \\
    \textbf{FoF} & \textbf{94.17 $\pm$ 0.68} & \textbf{89.98 $\pm$ 1.35} & \textbf{79.98 $\pm$ 2.17} & \textbf{69.05 $\pm$ 2.24} \\
\bottomrule
\end{tabular}
\end{table*}

\begin{table*}[t]
\setlength{\tabcolsep}{12pt}
\centering
\caption{Comparison with \textbf{multimodal grading} on the TCGA GBM-LGG dataset.}
\label{tab:multi}
\begin{tabular}{l|cccc}
\toprule
Method & AUC (\%) & AP (\%) & Accuracy (\%) & Kappa (\%) \\
\hline
    SCNN \cite{mobadersany2018predicting} & 91.18 $\pm$ 0.85 & 84.40 $\pm$ 1.83 & 74.40 $\pm$ 2.09 & 61.02 $\pm$ 1.34 \\
    Pathomic \cite{chen2020pathomic}  & 92.30 $\pm$ 0.79  & 85.80 $\pm$ 1.55 & 75.74 $\pm$ 1.39 & 63.15 $\pm$ 2.31 \\
    Gpdbn \cite{wang2021gpdbn}  & 92.01 $\pm$ 0.56  & 85.42 $\pm$ 1.12 & 75.98 $\pm$ 1.42 & 63.02 $\pm$ 1.56 \\
    Hfbsurv \cite{li2022hfbsurv}  & 91.61 $\pm$ 0.75  & 84.79 $\pm$ 1.39 & 74.67 $\pm$ 2.21 & 61.84 $\pm$ 1.65 \\
    DDM-net \cite{qiu2024dual}  & 92.53 $\pm$ 0.91  & 85.22 $\pm$ 1.01 & 76.92 $\pm$ 1.76 & 64.07 $\pm$ 1.95 \\
    FOAA \cite{alwazzan2024foaa}  & 93.15 $\pm$ 0.76  & 86.70 $\pm$ 1.14 & 77.90 $\pm$ 2.41 & 65.36 $\pm$ 2.38 \\
    DGKD \cite{xing2022discrepancy}  & 93.43 $\pm$ 0.40  & 87.75 $\pm$ 0.86 & 78.08 $\pm$ 0.73 & 66.73 $\pm$ 1.31 \\
    \hline
    \textbf{FoF} & \textbf{94.17 $\pm$ 0.68} & \textbf{89.98 $\pm$ 1.35} & \textbf{79.98 $\pm$ 2.17} & \textbf{69.05 $\pm$ 2.24} \\
\bottomrule
\end{tabular}
\end{table*}

\subsection{Training and Inference}
The overall optimization objective of our FoF framework is summarized as follows:
\begin{equation}
    \mathcal{L} \ = \ \mathcal{L}_{\rm{CLS}} + \lambda_1 \mathcal{L}_{\rm{FRL}} + \lambda_2 \mathcal{L}_{\rm{MCA}},
\end{equation}
where $\lambda_1$ and $\lambda_2$ indicate the coefficients to control the trade-off of FRL and MCA, respectively. During training, FoF estimates the contribution score $\mathbf{A} \in \mathbb{R}^{H\times W}$ in the first back-propagation and calculates the losses of FRL and MCA modules in the second back-propagation. At the inference phase, the model $f\cdot g$ directly outputs the predictions of the input images with no additional computational cost.

\section{Experiment}
\subsection{Dataset and Implementation Details}
\noindent\textbf{Dataset.} We follow \cite{xing2022discrepancy} to evaluate our FoF framework on the combination of TCGA-GBM and TCGA-LGG datasets \cite{tomczak2015review}, which comprise paired histopathology slides and genomic profiles. There are a total of $736$ patients with standard grading labels, including $182$ grade II, $205$ grade III, and $350$ grade IV. Following \cite{chen2020pathomic, mobadersany2018predicting}, we utilize $1,325$ Region of Interest (ROI) images of dimensions $1,024\times 1,024$ from pathology modality. The ROI images are augmented through random cropping, color jittering, flipping, and distortion. We perform 5-fold cross-validation and report the average test performance with the standard deviation.


\noindent\textbf{Implementation.} We implement the proposed framework with the PyTorch library \cite{paszke2019pytorch} and employ ViT-Tiny \cite{dosovitskiy2020image, steiner2021train} as the encoder for images. All the experiments are done on four NVIDIA GeForce GTX 1080 Ti GPUs with a batch size of 4. 
For network optimization, we employ AdamW optimizer, configuring it with $\beta_1 = 0.9$, $\beta_2 = 0.999$, and a weight decay of $0.01$. 
The initial learning rate is set as $1 \times 10^{-4}$ and follows a cosine decay schedule down to 0. The framework is trained for 70 epochs. Both the loss weights $\lambda_1$ and $\lambda_2$ are set as 0.2. The temperature parameter $\tau$ is set as 0.07.

\begin{figure}\centering
\subfigure[Input slide]{\label{a}\includegraphics[width=.32\linewidth]{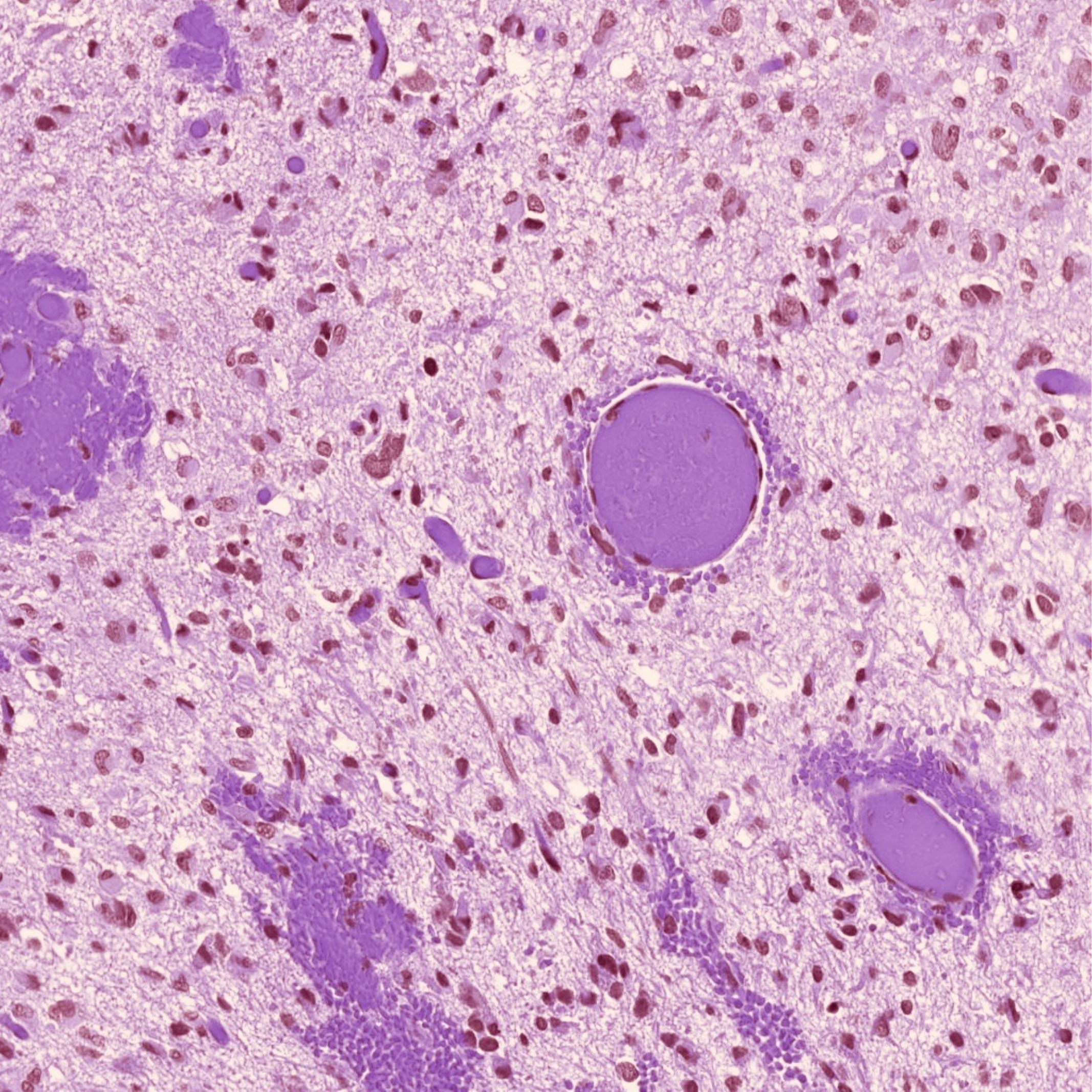}} 
\subfigure[Baseline]{\label{b}\includegraphics[width=.32\linewidth]{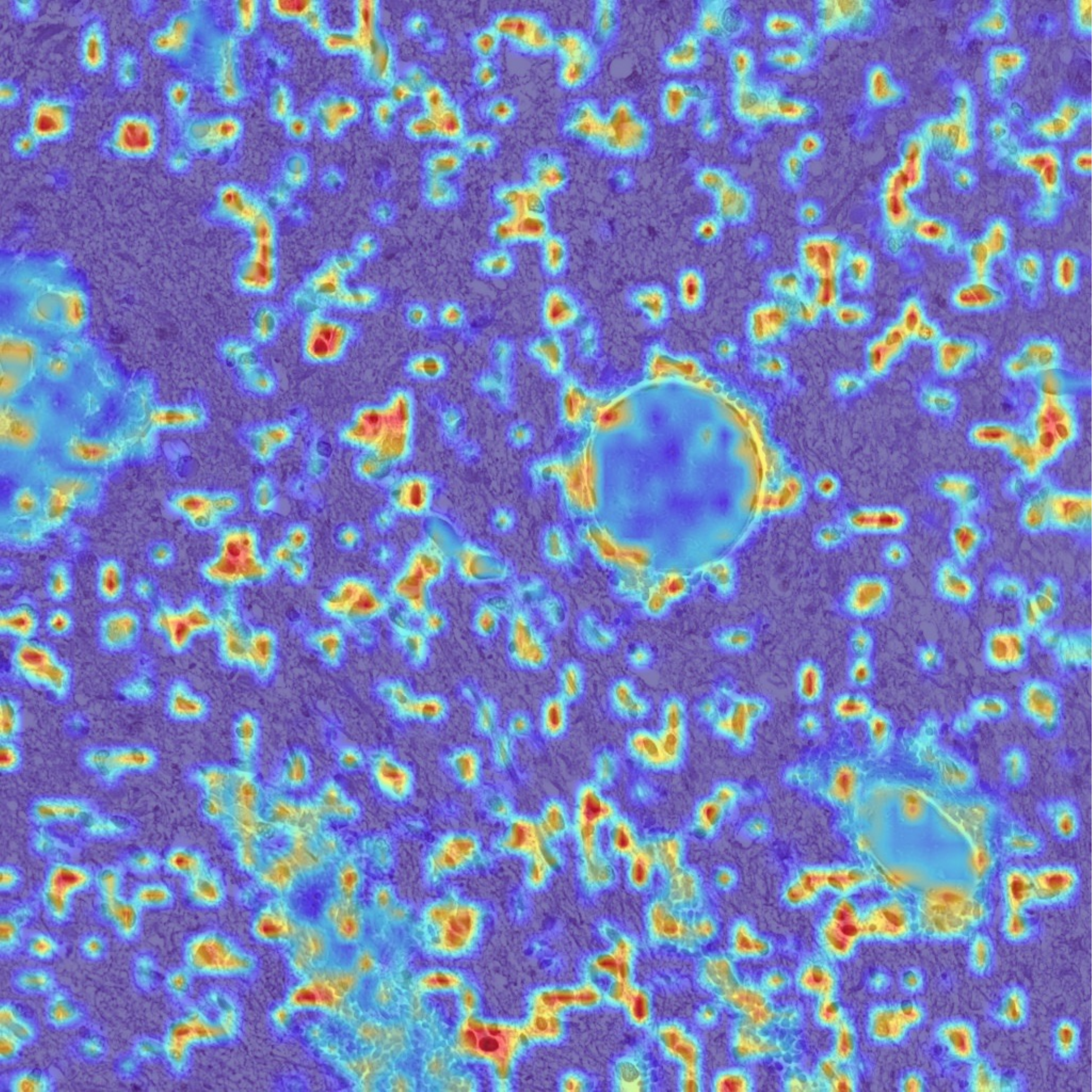}}
\subfigure[FoF]{\label{c}\includegraphics[width=.32\linewidth]{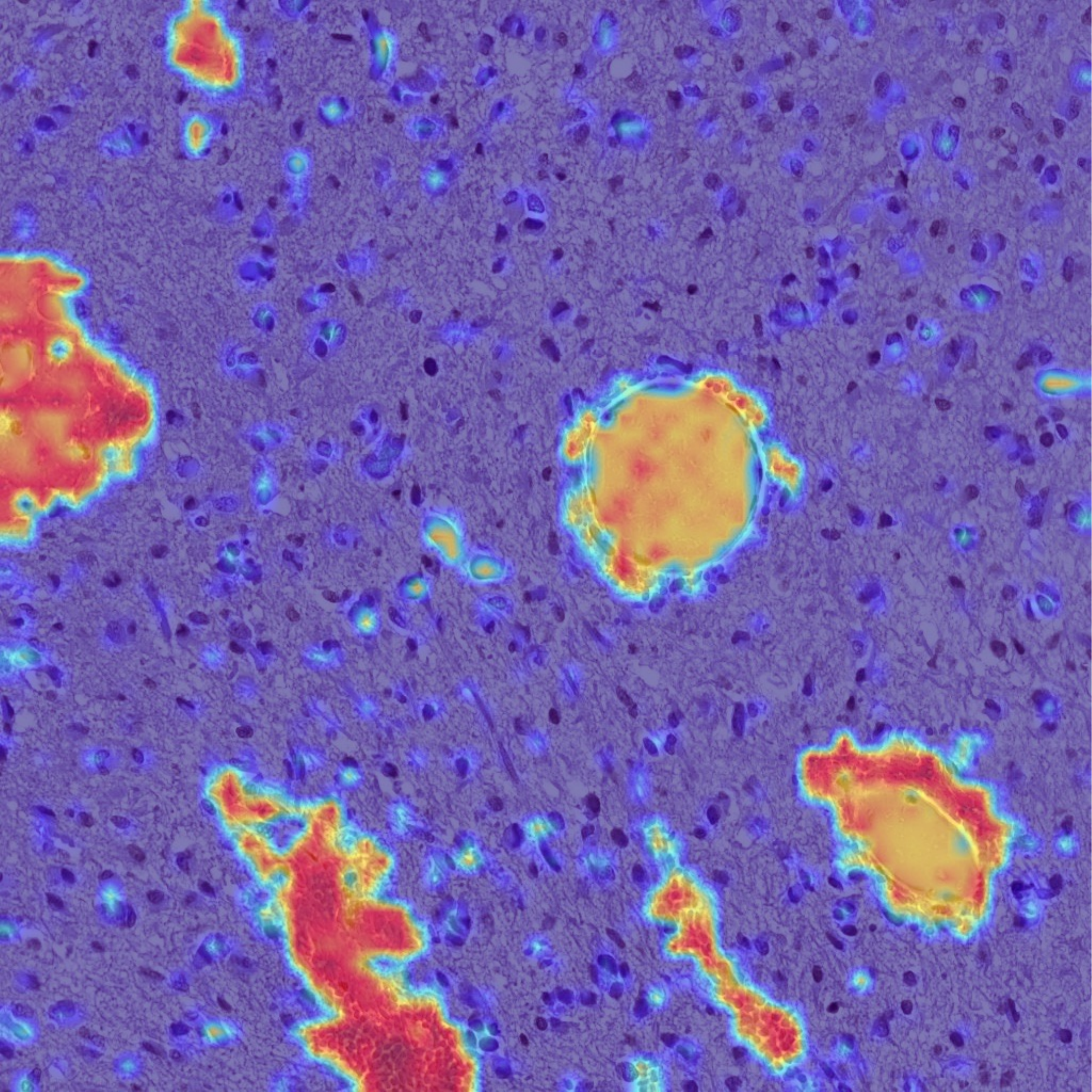}}
\caption{Visualization of (a) the input pathology slide, (b) the CAM produced by baseline method ViT-Tiny, and (c) the CAM generated by the proposed FoF framework. As illustrated, FoF focuses on microvascular proliferation, which leads to the diagnosis of Glioblastoma (Grade IV).}
\label{fig:example}
\end{figure}

\begin{figure*}[t!]
\centering
\subfigure[IDH]{
\includegraphics[width=.32\textwidth]{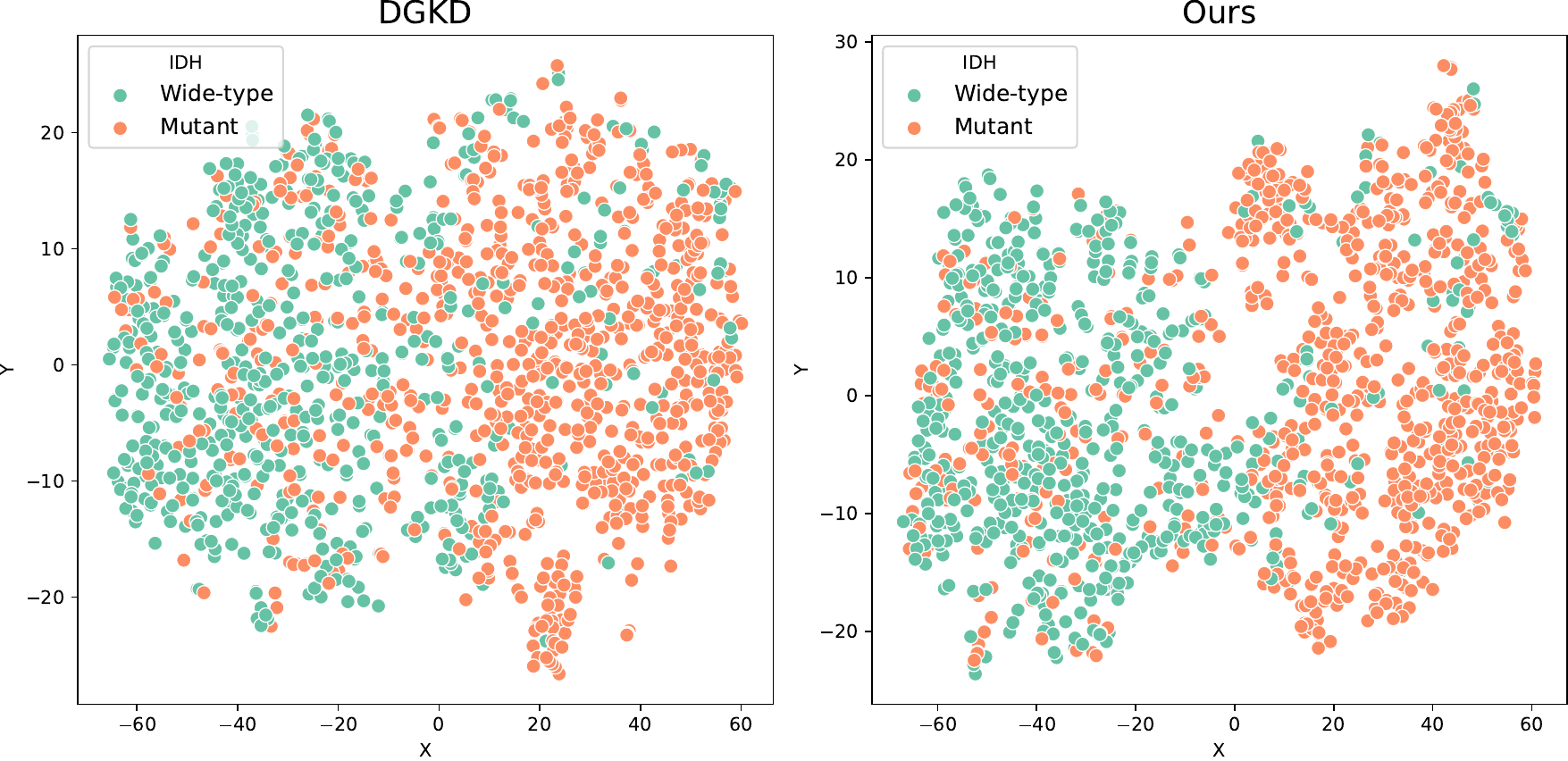}}
\subfigure[1p/19q]{
\includegraphics[width=.32\textwidth]{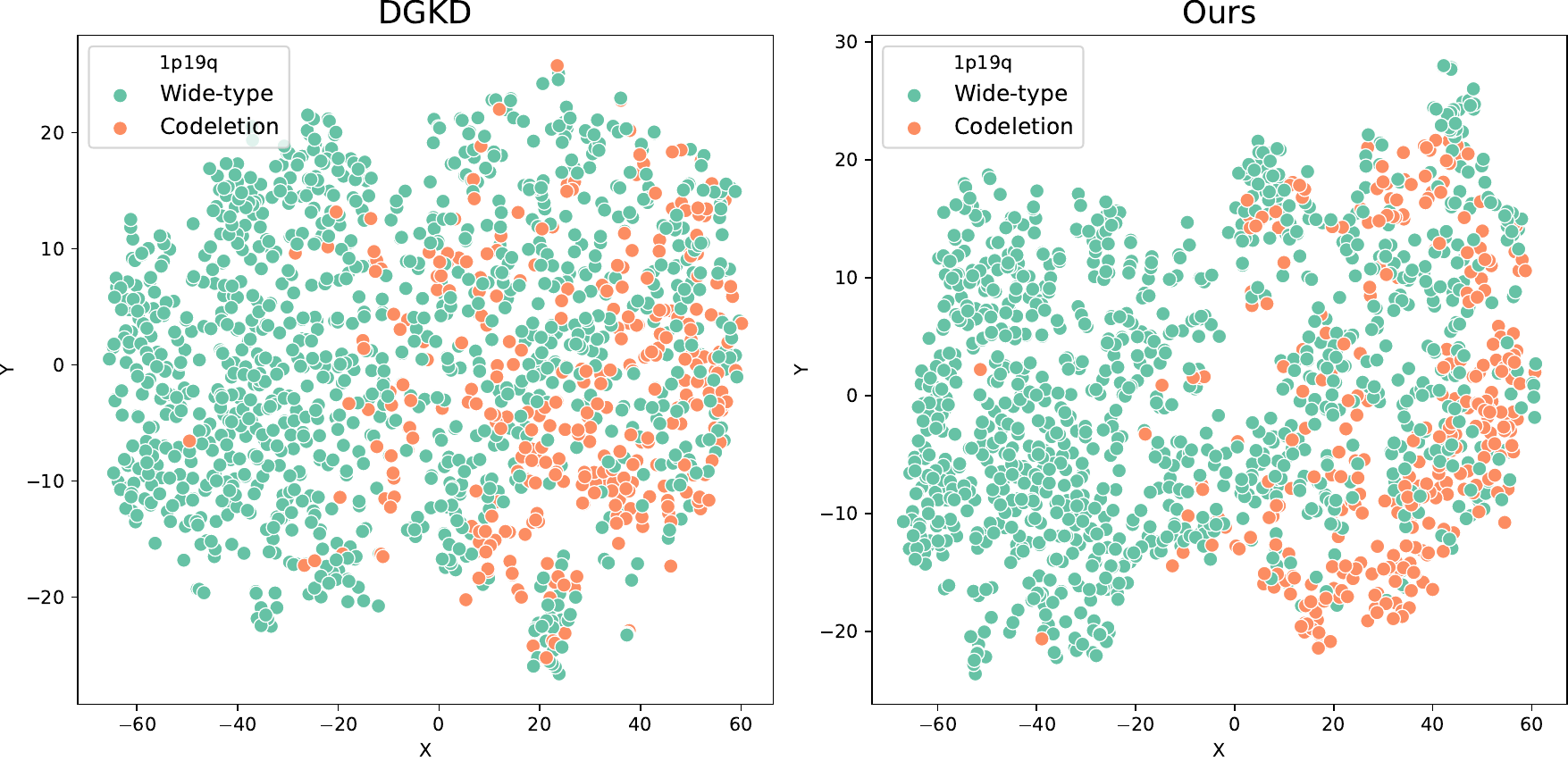}}
\subfigure[CARD11]{
\includegraphics[width=.32\textwidth]{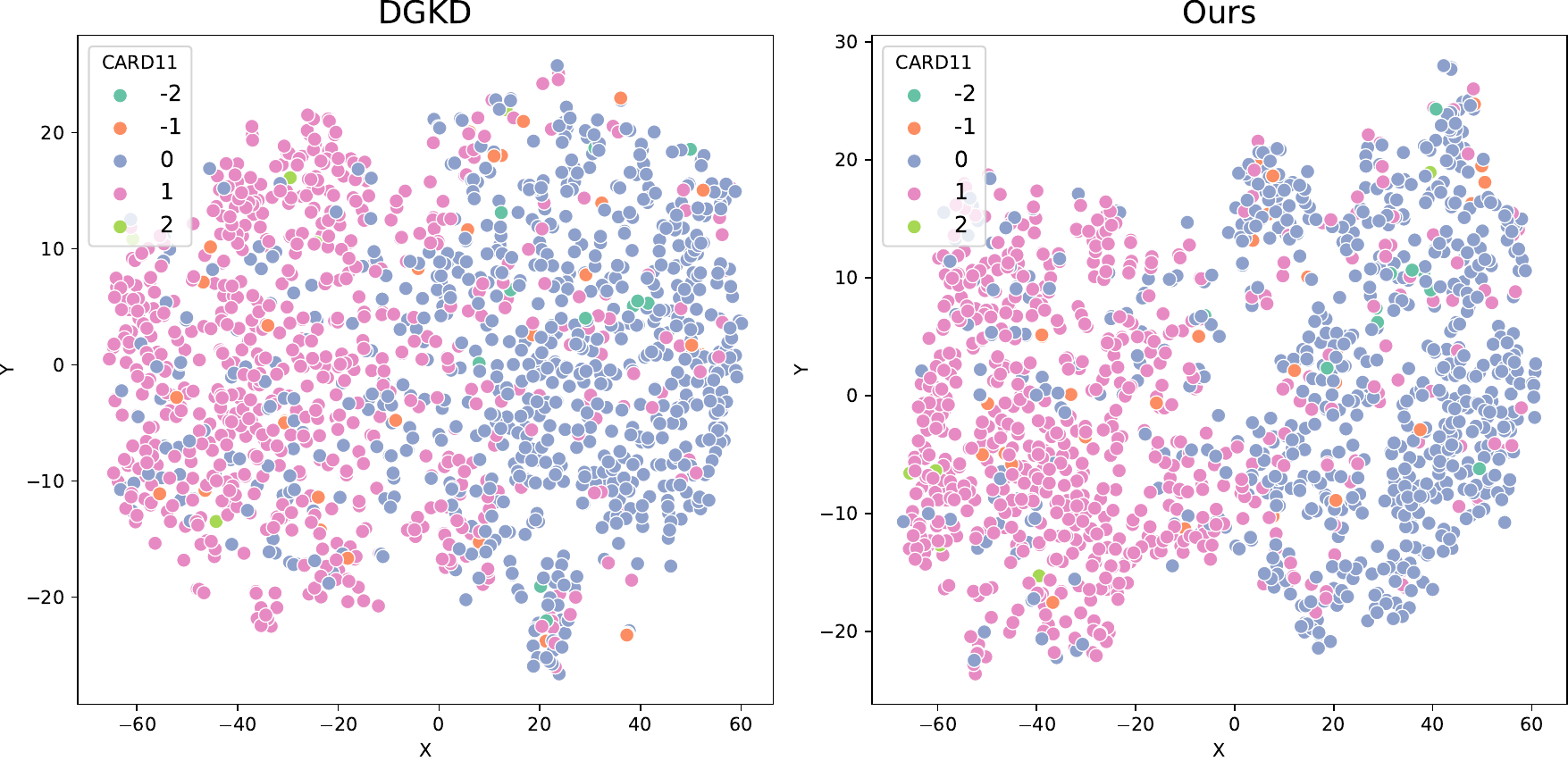}}
\subfigure[EGFR]{
\includegraphics[width=.32\textwidth]{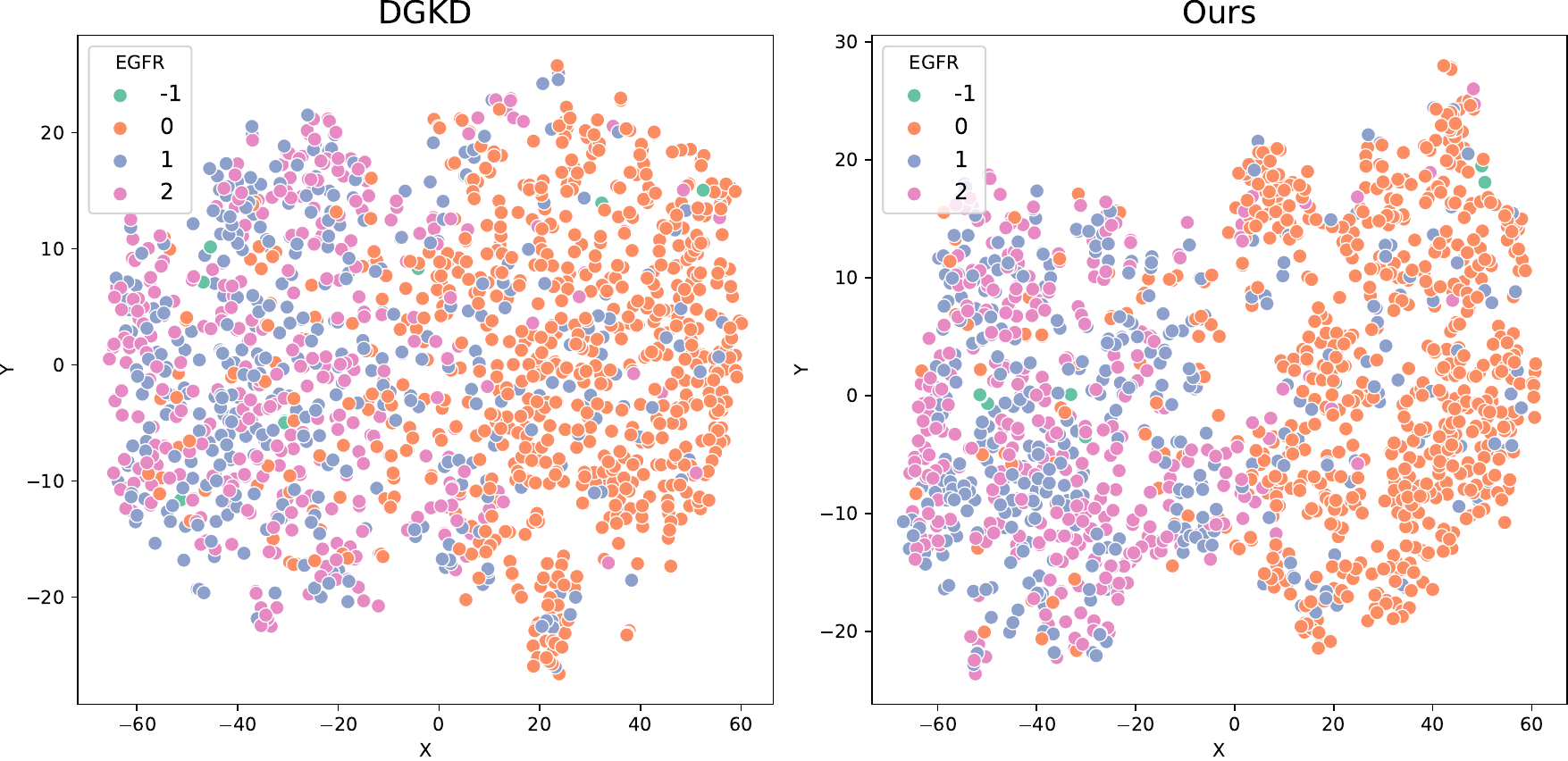}}
\subfigure[FGFR2]{
\includegraphics[width=.32\textwidth]{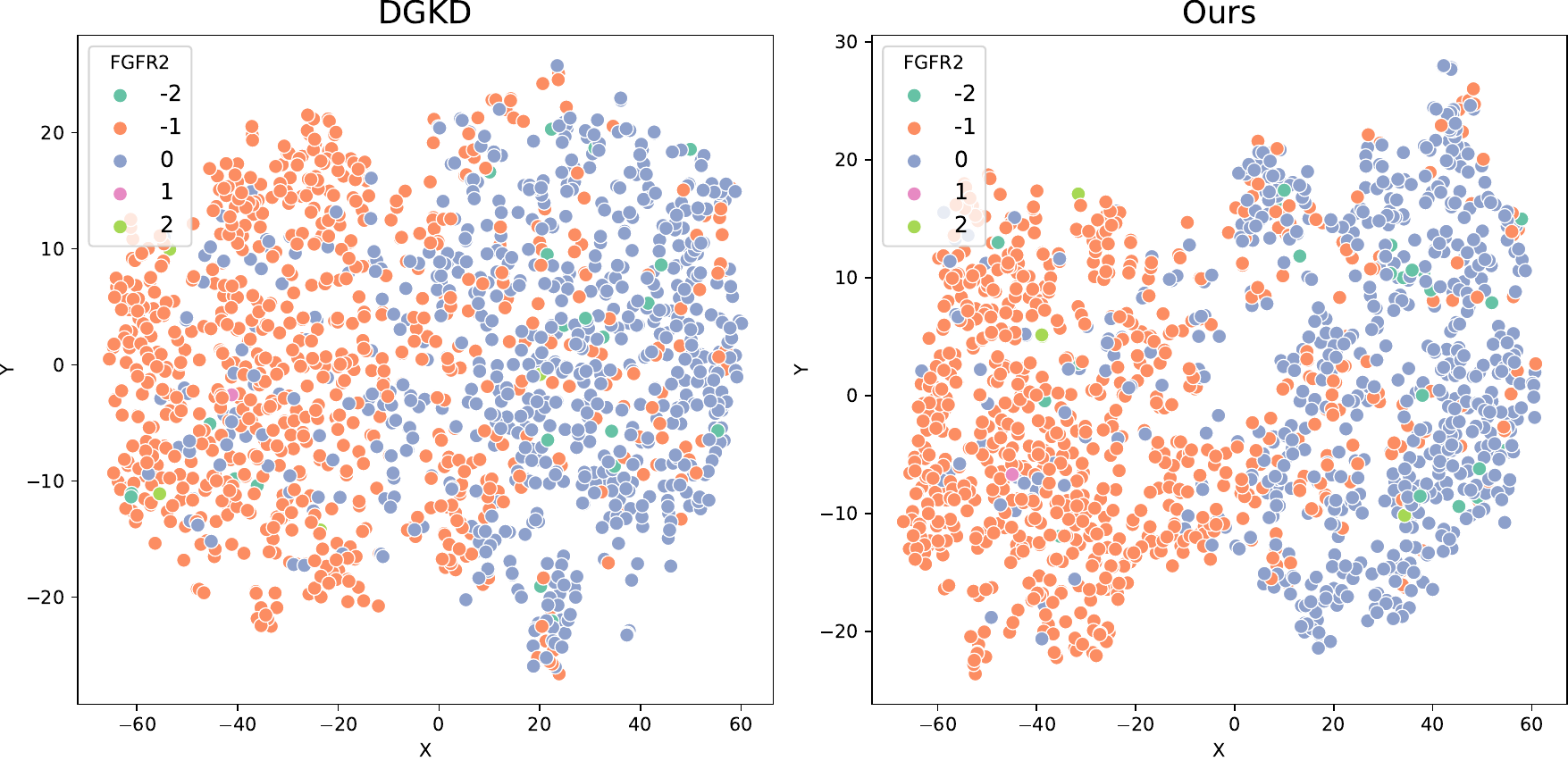}}
\subfigure[PTEN]{
\includegraphics[width=.32\textwidth]{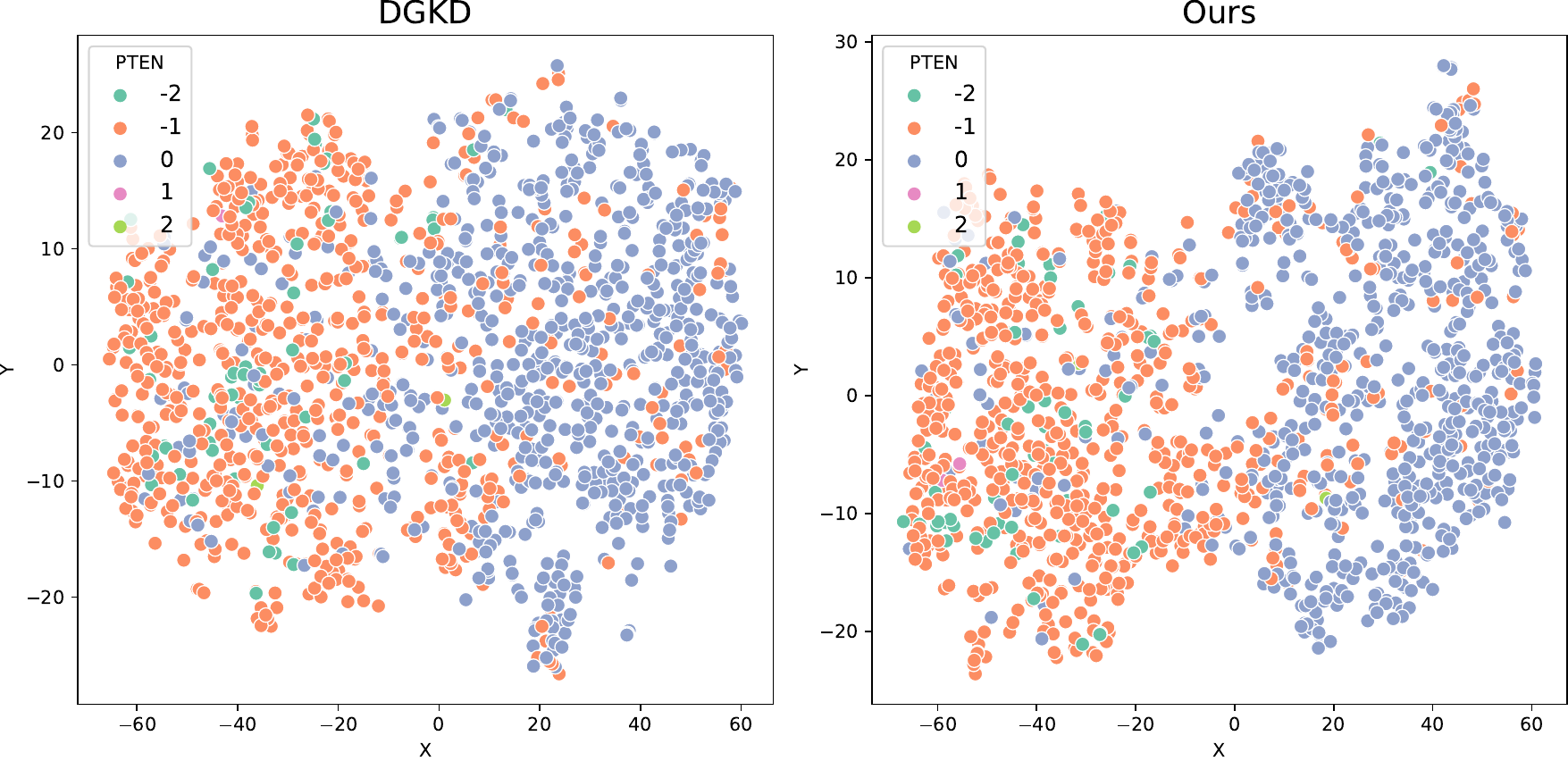}}
\caption{Comparisons with the baseline on the molecular-pathology knowledge alignment.}
\label{fig:tsne}
\end{figure*}

\begin{figure*}[t!]
\centering
\subfigure{
\includegraphics[width=.32\textwidth]{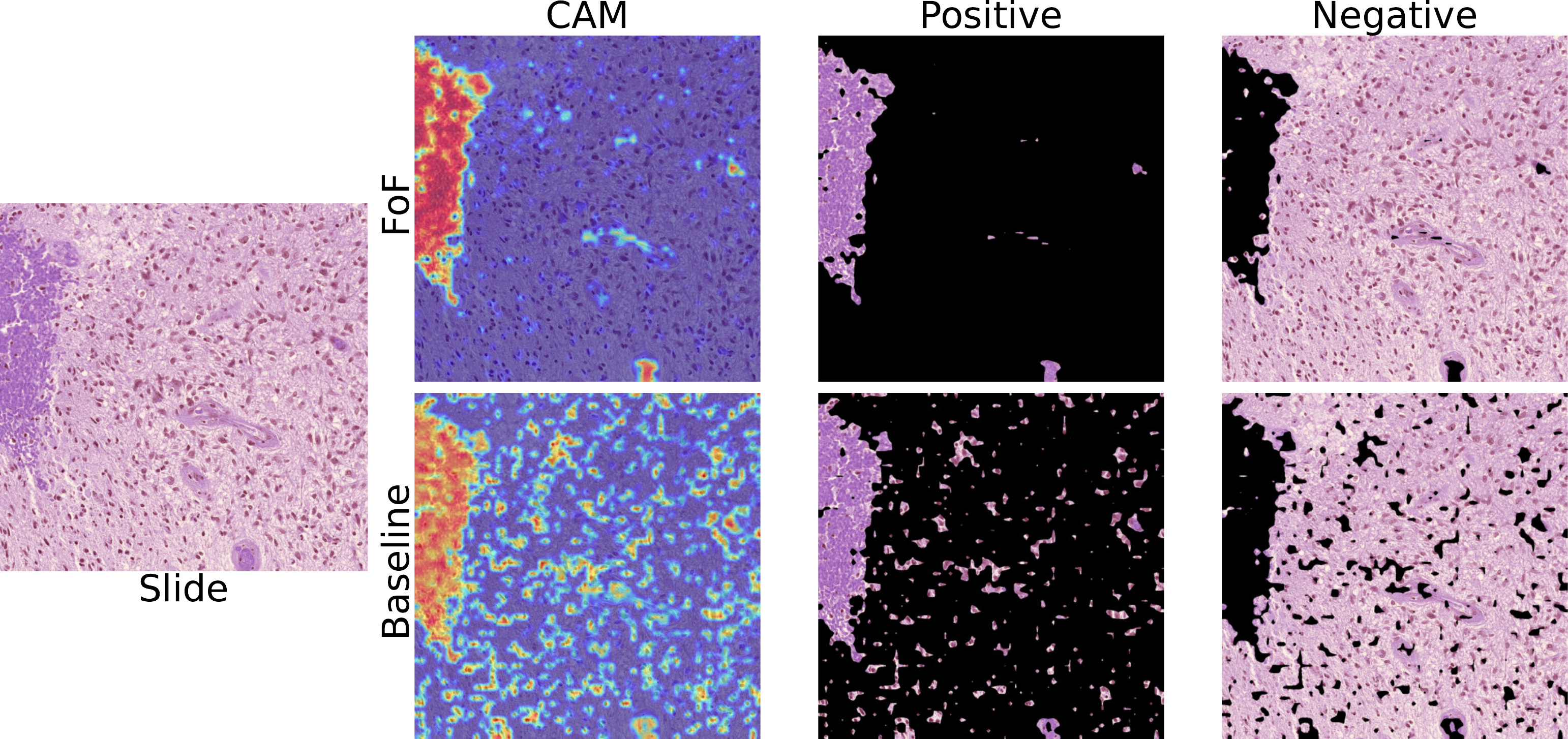}}
\subfigure{
\includegraphics[width=.32\textwidth]{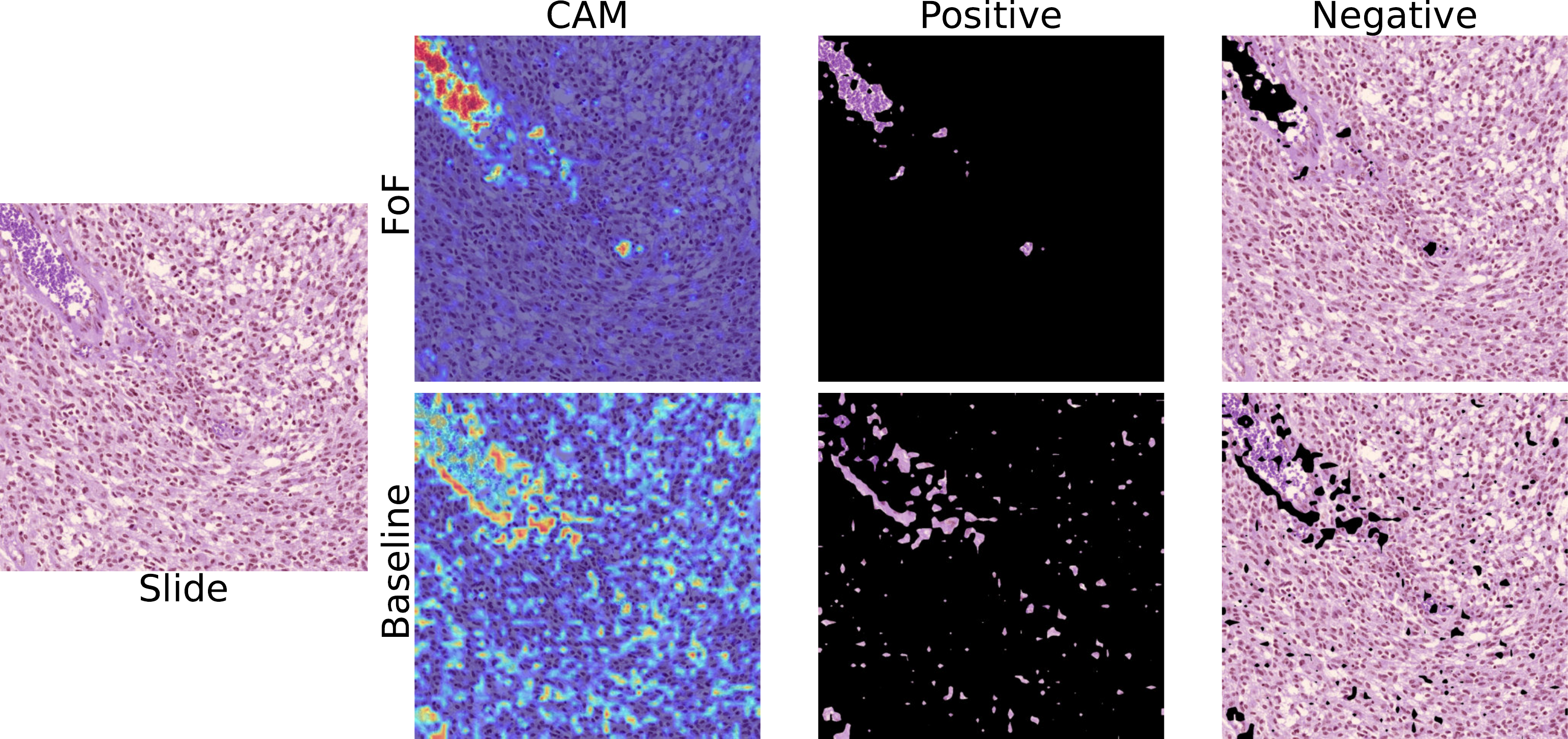}}
\subfigure{
\includegraphics[width=.32\textwidth]{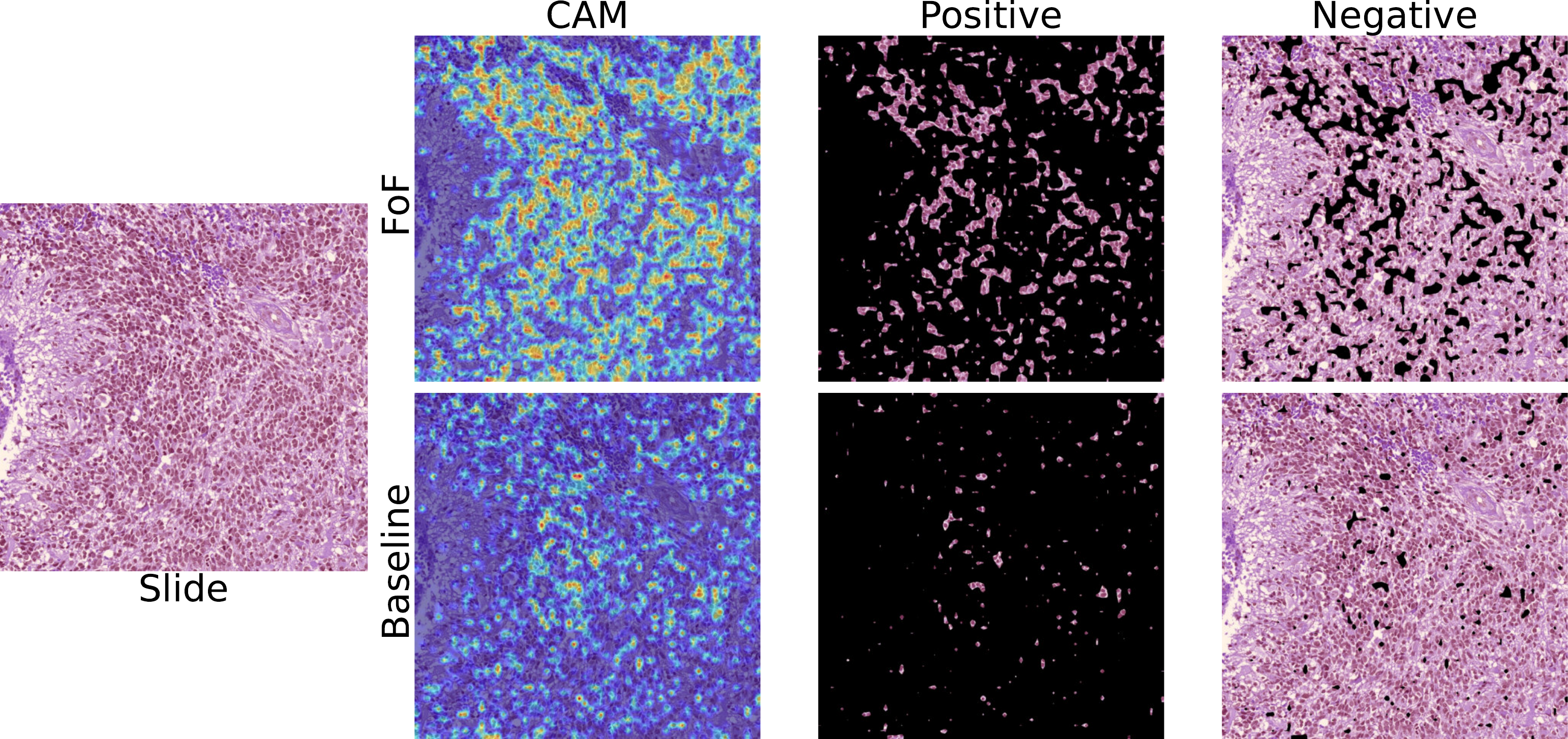}}
\subfigure{
\includegraphics[width=.32\textwidth]{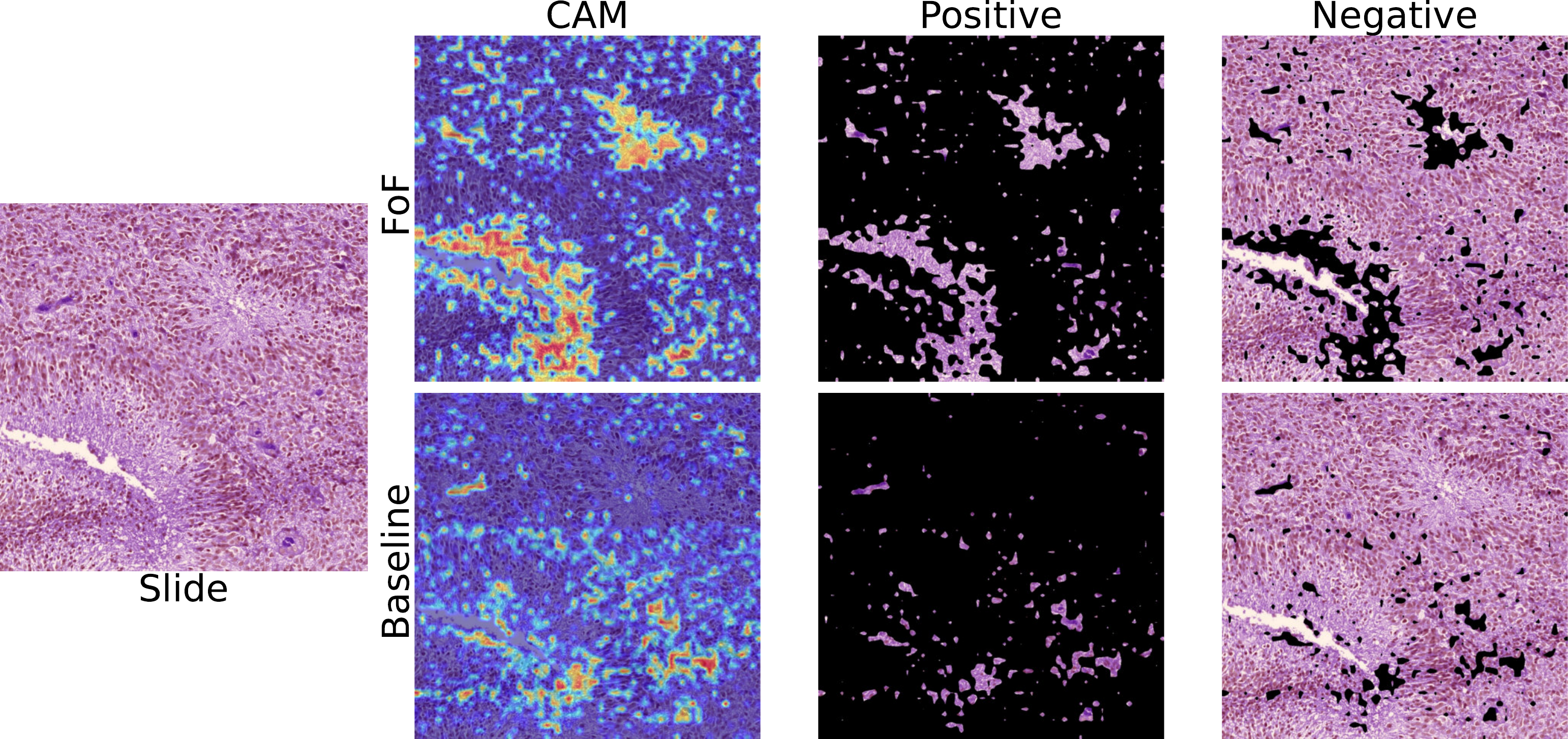}}
\subfigure{
\includegraphics[width=.32\textwidth]{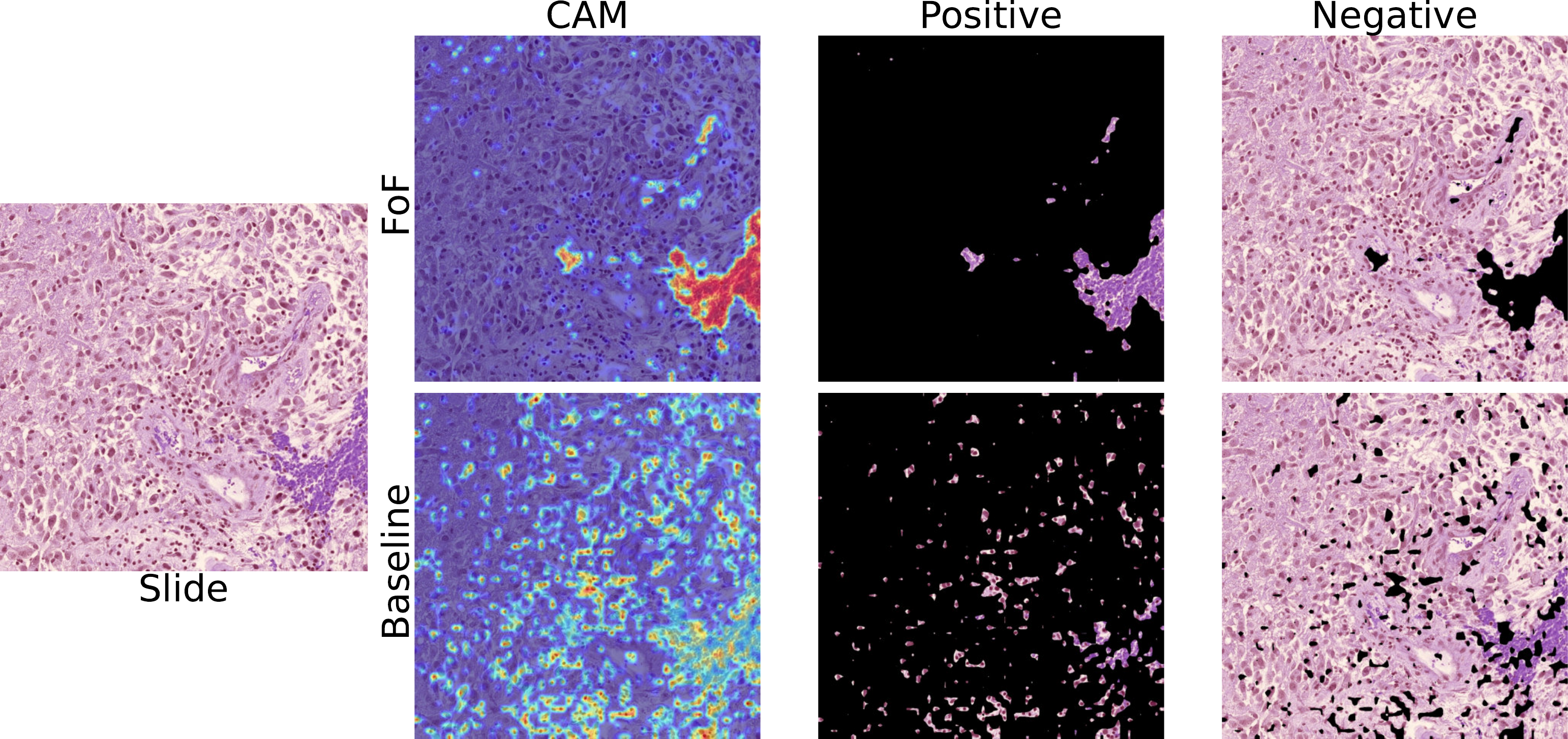}}
\subfigure{
\includegraphics[width=.32\textwidth]{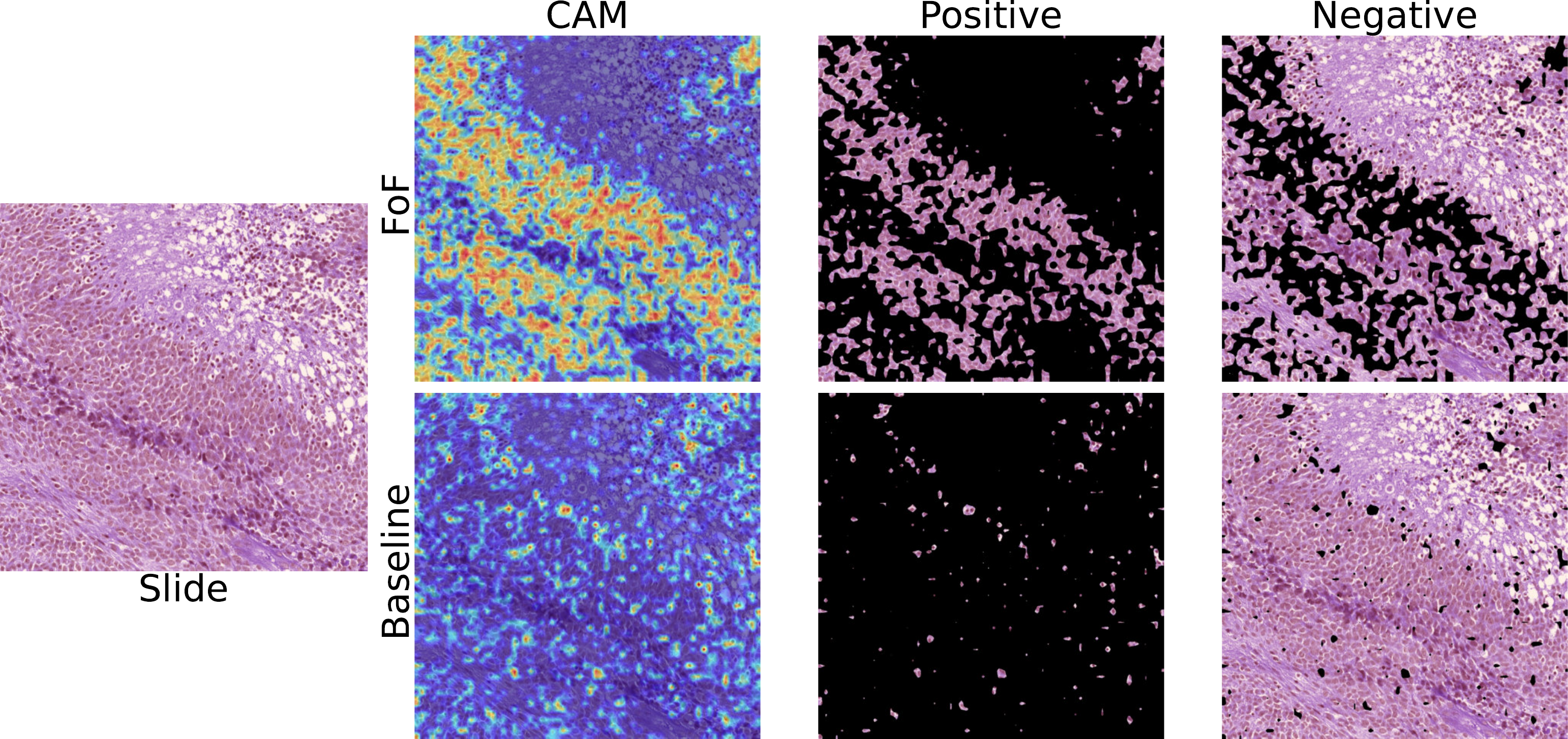}}
\caption{Comparison with the baseline on model focus. As illustrated, our FoF framework concentrates more precisely on the key structures of Glioblastoma, such as microvascular proliferation and pseudopalisadeing necrosis.}
\label{fig:cam}
\end{figure*}

\noindent\textbf{Evaluation Metrics.}
Following the evaluation protocol in \cite{xing2022discrepancy,xing2024comprehensive}, we adopt four commonly-used metrics to comprehensively evaluate the performance of glioma grading, including the Average Precision (AP), Accuracy, Area Under the Curve (AUC), and Kappa score. Higher scores of these metrics represent a more accurate classification of the gliomas.

\subsection{Comparison with Histopathology Grading Methods}
We compare our FoF with state-of-the-art methods that require only histopathology slides as input during inference, including popular image classification models (ViT-Tiny as Baseline, PathCNN \cite{xing2022discrepancy}, PathGCN \cite{chen2020pathomic}, and SwinV2 \cite{liu2022swin}). For a more fair comparison, we implement various latest relevant comparisons based on knowledge distillation, which also utilizes multimodal data for training while requiring image-only data for inference. As shown in Table \ref{tab:path}, FoF outperforms other histopathology grading methods in all metrics on the TCGA GBM-LGG dataset, achieving an Accuracy of 79.98\%, an AUC of 94.17\%, an Average Precision of 89.98\%, and a Kappa of 69.05\%. FoF significantly outperforms the baseline ViT model, showing 5.62\%, 3.80\%, 6.93\%, 7.89\% increases in Accuracy, AUC, Average Precision, Kappa, respectively.
Notably, compared with the most recent study CLAT \cite{xing2024comprehensive},
FoF achieves a 3.51\% increase in Accuracy, 1.75\% increase in AUC, 3.64\% increase in Average Precision, and 4.71\% increase in Kappa. These comparisons confirm the effectiveness of our FoF framework.

We have also compared the Class Activation Map generated by the baseline model and our FoF framework. As illustrated in Fig. \ref{fig:example}, our FoF framework focuses more precisely on regions of microvascular proliferation, a key histopathology feature of Glioblastoma (Grade IV). 

\noindent\textbf{Ablation study.} We perform the ablation study to validate the effectiveness of our proposed FRL and MCA modules on the TCGA GBM-LGG dataset. As shown in Table \ref{tab:path}, both FRL and MCA modules remarkably improve the performance over the baselines. In particular, we apply two ablative baselines of the proposed FoF framework by disabling the FRL (denoted as FoF $w/o$ FRL) and the MCA (denoted as FoF $w/o$ MCA) individually. In detail, as shown in Table \ref{tab:path}, for FRL, the Accuracy increases by 3.09\%, AUC increases by 0.97\%, Average Precision increases by 1.54\%, and Kappa increases by 4.74\%. For MCA, the Accuracy increases by 3.43\%, AUC increases by 0.53\%, Average Precision increases by 1.72\%, and Kappa increases by 5.20\%. The ablation study verifies the effectiveness of our FRL and MCA modules.

\subsection{Comparison with Multimodal Grading Methods} 
We compare our FoF with multimodal grading methods that require paired histopathology slides and genomic data for glioma grading. These methods implement various feature fusion methods, including concatenation (SCNN \cite{mobadersany2018predicting}, Gpdbn \cite{wang2021gpdbn}), Kronecker Product (Pathomic \cite{chen2020pathomic}, DGKD \cite{xing2022discrepancy}), hierarchical fusion (Hfbsurv \cite{li2022hfbsurv}), and attention (DDM-net \cite{qiu2024dual}, FOAA \cite{alwazzan2024foaa}).
As shown in Table \ref{tab:multi}, our FoF with sole histopathology slides outperforms all the multimodal grading methods on all metrics, significantly illustrating the effectiveness of the proposed FoF framework. In detail, our FoF outperforms the most recent multimodal grading method DGKD \cite{xing2022discrepancy} with improvements of 1.9\% in Accuracy, 0.74\% in AUC, 2.23\% in Average Precision, and 2.32\% in Kappa. Compared to FOAA \cite{alwazzan2024foaa}, which employs more advanced attention-based feature fusion, FoF achieved 2.08\%, 1.02\%, 3.28\%, 3.69\% increases in Accuracy, AUC, Average Precision, Kappa, respectively.

\subsection{Qualitative Analysis}
We have illustrated the distributions of morphological features regarding the different values of molecular biomarkers in Fig. \ref{fig:tsne}. Compared to DGKD \cite{xing2022discrepancy}, our FoF presents a clearer clustering of histopathology features with respect to the molecular biomarkers, illustrating a more effective alignment between the two modalities. 

Furthermore, we present more comparisons of the model focus on input histopathology slides in Fig. \ref{fig:cam}. All the images presented have been reviewed by senior pathologists. It is worth noting that our FoF concentrates on both microvascular proliferation and pseudopalisading necrosis, critical indicators for diagnosing HGG, whereas the baseline model focuses on other, less relevant details.

\section{Conclusion}
In this work, we propose the FoF framework that utilizes pathology-genomic knowledge toward accurate glioma grading with histopathology slides. To enhance the representation learning on histopathology slides, we propose the FRL module that encourages the model to focus on diagnostic regions and the MCA scheme which efficiently aligns the molecular biomarkers with visual representations. Experimental results indicate that FoF improves the glioma grading significantly. Particularly, with the sole histopathology slides, FoF achieves superior performance with existing multimodal approaches, which is of great clinical significance.
\balance
\bibliographystyle{IEEEtran}
\bibliography{refs}

\begin{thebibliography}{10}
\providecommand{\url}[1]{#1}
\csname url@samestyle\endcsname
\providecommand{\newblock}{\relax}
\providecommand{\bibinfo}[2]{#2}
\providecommand{\BIBentrySTDinterwordspacing}{\spaceskip=0pt\relax}
\providecommand{\BIBentryALTinterwordstretchfactor}{4}
\providecommand{\BIBentryALTinterwordspacing}{\spaceskip=\fontdimen2\font plus
\BIBentryALTinterwordstretchfactor\fontdimen3\font minus \fontdimen4\font\relax}
\providecommand{\BIBforeignlanguage}[2]{{%
\expandafter\ifx\csname l@#1\endcsname\relax
\typeout{** WARNING: IEEEtran.bst: No hyphenation pattern has been}%
\typeout{** loaded for the language `#1'. Using the pattern for}%
\typeout{** the default language instead.}%
\else
\language=\csname l@#1\endcsname
\fi
#2}}
\providecommand{\BIBdecl}{\relax}
\BIBdecl

\bibitem{louis20212021}
D.~N. Louis, A.~Perry, P.~Wesseling, D.~J. Brat, I.~A. Cree, D.~Figarella-Branger, C.~Hawkins, H.~Ng, S.~M. Pfister, G.~Reifenberger \emph{et~al.}, ``The 2021 who classification of tumors of the central nervous system: a summary,'' \emph{Neuro-oncology}, vol.~23, no.~8, pp. 1231--1251, 2021.

\bibitem{louis20162016}
D.~N. Louis, A.~Perry, G.~Reifenberger, A.~Von~Deimling, D.~Figarella-Branger, W.~K. Cavenee, H.~Ohgaki, O.~D. Wiestler, P.~Kleihues, and D.~W. Ellison, ``The 2016 world health organization classification of tumors of the central nervous system: a summary,'' \emph{Acta neuropathologica}, vol. 131, pp. 803--820, 2016.

\bibitem{shanes1987interobserver}
J.~Shanes, J.~Ghali, M.~Billingham, V.~Ferrans, J.~Fenoglio, W.~Edwards, C.~Tsai, J.~Saffitz, J.~Isner, and S.~Furner, ``Interobserver variability in the pathologic interpretation of endomyocardial biopsy results.'' \emph{Circulation}, vol.~75, no.~2, pp. 401--405, 1987.

\bibitem{anand2020histographs}
D.~Anand, S.~Gadiya, and A.~Sethi, ``Histographs: graphs in histopathology,'' in \emph{Medical Imaging 2020: Digital Pathology}, vol. 11320.\hskip 1em plus 0.5em minus 0.4em\relax SPIE, 2020, pp. 150--155.

\bibitem{wang2020weakly}
J.~Wang, R.~J. Chen, M.~Y. Lu, A.~Baras, and F.~Mahmood, ``Weakly supervised prostate tma classification via graph convolutional networks,'' in \emph{ISBI}.\hskip 1em plus 0.5em minus 0.4em\relax IEEE, 2020, pp. 239--243.

\bibitem{chen2020pathomic}
R.~J. Chen, M.~Y. Lu, J.~Wang, D.~F. Williamson, S.~J. Rodig, N.~I. Lindeman, and F.~Mahmood, ``Pathomic fusion: An integrated framework for fusing histopathology and genomic features for cancer diagnosis and prognosis,'' \emph{IEEE Transactions on Medical Imaging}, 2020.

\bibitem{mobadersany2018predicting}
P.~Mobadersany, S.~Yousefi, M.~Amgad, D.~A. Gutman, J.~S. Barnholtz-Sloan, J.~E. Vel{\'a}zquez~Vega, D.~J. Brat, and L.~A. Cooper, ``Predicting cancer outcomes from histology and genomics using convolutional networks,'' \emph{Proceedings of the National Academy of Sciences}, vol. 115, no.~13, pp. E2970--E2979, 2018.

\bibitem{alwazzan2024foaa}
O.~Alwazzan, I.~Patras, and G.~Slabaugh, ``Foaa: Flattened outer arithmetic attention for multimodal tumor classification,'' \emph{arXiv preprint arXiv:2403.06339}, 2024.

\bibitem{zhou2021distilling}
S.~Zhou, Y.~Wang, D.~Chen, J.~Chen, X.~Wang, C.~Wang, and J.~Bu, ``Distilling holistic knowledge with graph neural networks,'' in \emph{ICCV}, 2021, pp. 10\,387--10\,396.

\bibitem{xing2024comprehensive}
X.~Xing, M.~Zhu, Z.~Chen, and Y.~Yuan, ``Comprehensive learning and adaptive teaching: Distilling multi-modal knowledge for pathological glioma grading,'' \emph{Medical Image Analysis}, vol.~91, p. 102990, 2024.

\bibitem{zhang2023multi}
S.~Zhang, S.~Zheng, Z.~Shui, H.~Li, and L.~Yang, ``Multi-modal learning with missing modality in predicting axillary lymph node metastasis,'' in \emph{2023 IEEE International Conference on Bioinformatics and Biomedicine (BIBM)}.\hskip 1em plus 0.5em minus 0.4em\relax IEEE, 2023, pp. 2395--2400.

\bibitem{wang2019machine}
X.~Wang, D.~Wang, Z.~Yao, B.~Xin, B.~Wang, C.~Lan, Y.~Qin, S.~Xu, D.~He, and Y.~Liu, ``Machine learning models for multiparametric glioma grading with quantitative result interpretations,'' \emph{Frontiers in neuroscience}, vol.~12, p. 1046, 2019.

\bibitem{zhou2019cgc}
Y.~Zhou, S.~Graham, N.~Alemi~Koohbanani, M.~Shaban, P.-A. Heng, and N.~Rajpoot, ``Cgc-net: Cell graph convolutional network for grading of colorectal cancer histology images,'' in \emph{ICCV workshops}, 2019, pp. 0--0.

\bibitem{wang2023multi}
X.~Wang, S.~Price, and C.~Li, ``Multi-task learning of histology and molecular markers for classifying diffuse glioma,'' \emph{arXiv preprint arXiv:2303.14845}, 2023.

\bibitem{perry2016histologic}
A.~Perry and P.~Wesseling, ``Histologic classification of gliomas,'' \emph{Handbook of clinical neurology}, vol. 134, pp. 71--95, 2016.

\bibitem{ker2019automated}
J.~Ker, Y.~Bai, H.~Y. Lee, J.~Rao, and L.~Wang, ``Automated brain histology classification using machine learning,'' \emph{Journal of Clinical Neuroscience}, vol.~66, pp. 239--245, 2019.

\bibitem{rathore2020glioma}
S.~Rathore, T.~Niazi, M.~A. Iftikhar, and A.~Chaddad, ``Glioma grading via analysis of digital pathology images using machine learning,'' \emph{Cancers}, vol.~12, no.~3, p. 578, 2020.

\bibitem{wang2021gpdbn}
Z.~Wang, R.~Li, M.~Wang, and A.~Li, ``Gpdbn: deep bilinear network integrating both genomic data and pathological images for breast cancer prognosis prediction,'' \emph{Bioinformatics}, vol.~37, no.~18, pp. 2963--2970, 2021.

\bibitem{xing2022discrepancy}
X.~Xing, Z.~Chen, M.~Zhu, Y.~Hou, Z.~Gao, and Y.~Yuan, ``Discrepancy and gradient-guided multi-modal knowledge distillation for pathological glioma grading,'' in \emph{MICCAI}.\hskip 1em plus 0.5em minus 0.4em\relax Springer, 2022, pp. 636--646.

\bibitem{xing2023gradient}
X.~Xing, Z.~Chen, Y.~Hou, and Y.~Yuan, ``Gradient modulated contrastive distillation of low-rank multi-modal knowledge for disease diagnosis,'' \emph{Medical Image Analysis}, vol.~88, p. 102874, 2023.

\bibitem{qiu2024dual}
L.~Qiu, L.~Zhao, W.~Zhao, and J.~Zhao, ``Dual-space disentangled-multimodal network (ddm-net) for glioma diagnosis and prognosis with incomplete pathology and genomic data,'' \emph{Physics in Medicine \& Biology}, vol.~69, no.~8, p. 085028, 2024.

\bibitem{selvaraju2017grad}
R.~R. Selvaraju, M.~Cogswell, A.~Das, R.~Vedantam, D.~Parikh, and D.~Batra, ``Grad-cam: Visual explanations from deep networks via gradient-based localization,'' in \emph{ICCV}, 2017, pp. 618--626.

\bibitem{chen2020simple}
T.~Chen, S.~Kornblith, M.~Norouzi, and G.~Hinton, ``A simple framework for contrastive learning of visual representations,'' in \emph{International conference on machine learning}.\hskip 1em plus 0.5em minus 0.4em\relax PMLR, 2020, pp. 1597--1607.

\bibitem{sledzinska2021prognostic}
P.~{\'S}ledzi{\'n}ska, M.~G. Bebyn, J.~Furtak, J.~Kowalewski, and M.~A. Lewandowska, ``Prognostic and predictive biomarkers in gliomas,'' \emph{International journal of molecular sciences}, vol.~22, no.~19, p. 10373, 2021.

\bibitem{liu2022swin}
Z.~Liu, H.~Hu, Y.~Lin, Z.~Yao, Z.~Xie, Y.~Wei, J.~Ning, Y.~Cao, Z.~Zhang, L.~Dong \emph{et~al.}, ``Swin transformer v2: Scaling up capacity and resolution,'' in \emph{CVPR}, 2022, pp. 12\,009--12\,019.

\bibitem{hinton2015distilling}
G.~Hinton, O.~Vinyals, and J.~Dean, ``Distilling the knowledge in a neural network,'' \emph{arXiv preprint arXiv:1503.02531}, 2015.

\bibitem{passalis2018learning}
N.~Passalis and A.~Tefas, ``Learning deep representations with probabilistic knowledge transfer,'' in \emph{ECCV}, 2018, pp. 268--284.

\bibitem{hu2020knowledge}
M.~Hu, M.~Maillard, Y.~Zhang, T.~Ciceri, G.~La~Barbera, I.~Bloch, and P.~Gori, ``Knowledge distillation from multi-modal to mono-modal segmentation networks,'' in \emph{MICCAI}.\hskip 1em plus 0.5em minus 0.4em\relax Springer, 2020, pp. 772--781.

\bibitem{tung2019similarity}
F.~Tung and G.~Mori, ``Similarity-preserving knowledge distillation,'' in \emph{ICCV}, 2019, pp. 1365--1374.

\bibitem{park2019relational}
W.~Park, D.~Kim, Y.~Lu, and M.~Cho, ``Relational knowledge distillation,'' in \emph{CVPR}, 2019, pp. 3967--3976.

\bibitem{tian2019contrastive}
Y.~Tian, D.~Krishnan, and P.~Isola, ``Contrastive representation distillation,'' \emph{arXiv preprint arXiv:1910.10699}, 2019.

\bibitem{zhu2021student}
Y.~Zhu and Y.~Wang, ``Student customized knowledge distillation: Bridging the gap between student and teacher,'' in \emph{ICCV}, 2021, pp. 5057--5066.

\bibitem{li2022hfbsurv}
R.~Li, X.~Wu, A.~Li, and M.~Wang, ``Hfbsurv: hierarchical multimodal fusion with factorized bilinear models for cancer survival prediction,'' \emph{Bioinformatics}, vol.~38, no.~9, pp. 2587--2594, 2022.

\bibitem{tomczak2015review}
K.~Tomczak, P.~Czerwi{\'n}ska, and M.~Wiznerowicz, ``Review the cancer genome atlas (tcga): an immeasurable source of knowledge,'' \emph{Contemporary Oncology/Wsp{\'o}{\l}czesna Onkologia}, vol. 2015, no.~1, pp. 68--77, 2015.

\bibitem{paszke2019pytorch}
A.~Paszke, S.~Gross, F.~Massa, A.~Lerer, J.~Bradbury, G.~Chanan, T.~Killeen, Z.~Lin, N.~Gimelshein, L.~Antiga \emph{et~al.}, ``Pytorch: An imperative style, high-performance deep learning library,'' \emph{Advances in neural information processing systems}, vol.~32, 2019.

\bibitem{dosovitskiy2020image}
A.~Dosovitskiy, L.~Beyer, A.~Kolesnikov, D.~Weissenborn, X.~Zhai, T.~Unterthiner, M.~Dehghani, M.~Minderer, G.~Heigold, S.~Gelly \emph{et~al.}, ``An image is worth 16x16 words: Transformers for image recognition at scale,'' \emph{arXiv preprint arXiv:2010.11929}, 2020.

\bibitem{steiner2021train}
A.~Steiner, A.~Kolesnikov, X.~Zhai, R.~Wightman, J.~Uszkoreit, and L.~Beyer, ``How to train your vit? data, augmentation, and regularization in vision transformers,'' \emph{arXiv preprint arXiv:2106.10270}, 2021.

\end{thebibliography}

\end{document}